\documentclass[11pt,lettersize,pdflatex]{article}
\pdfoutput=1

\usepackage{fullpage}
\usepackage{amsmath}
\usepackage{amssymb}
\usepackage{graphicx} 
\usepackage{color}

\usepackage[T1]{fontenc}
\usepackage[latin1]{inputenc}
\usepackage{float}
\newcommand{\sgn}{\mathrm{sgn}}

\newcommand{\vw}{\textsf{Vowpal Wabbit} }
\newcommand{\citep}[1]{\cite{#1}}
\newcommand{\bigads}{Big\_Ads }
\newcommand{\E}{{\mathbf{E}}}

\providecommand{\tabularnewline}{\\}
\floatstyle{ruled}
\newfloat{algorithm}{tbp}{loa}
\floatname{algorithm}{Algorithm}

\newtheorem{thm}{Theorem}[section]
\newtheorem{cor}{Corollary}[section]
\newtheorem{lemma}{Lemma}[section]
\newtheorem{assumption}{Assumption}[section]

\newenvironment{proof}{\noindent\textsc{Proof.} }{\hfill $\Box$}

\begin{document}

\title{Sparse Online Learning via Truncated Gradient}

\author{ 
{\bf John Langford}  \\ 
Yahoo Research          \\ 
jl@yahoo-inc.com
\and 
{\bf Lihong Li} \\
Rutgers Computer Science Department \\
lihong@cs.rutgers.edu
\and 
{\bf Tong Zhang}   \\ 
Rutgers Statistics Department \\          
tongz@rci.rutgers.edu
} 
\date{}

\maketitle

\begin{abstract}
We propose a general method called truncated gradient
to induce sparsity in the weights of
online learning algorithms with convex loss functions.  This
method has several essential properties:
\begin{enumerate}
\item The degree of sparsity is
continuous---a parameter controls the rate of sparsification from
no sparsification to total sparsification.
\item The approach is
theoretically motivated, and an instance of it
can be regarded as an online counterpart
of the popular $L_1$-regularization method in the batch setting.
We prove that small rates of
sparsification result in only small additional regret with respect
to typical online learning guarantees.
\item The approach works well empirically.
\end{enumerate}
We apply the approach to several datasets and find that for
datasets with large numbers of features, substantial sparsity is
discoverable.
\end{abstract}

\section{Introduction}

We are concerned with machine learning over large datasets.  As an
example, the largest dataset we use here has over $10^7$ sparse
examples and $10^9$ features using about $10^{11}$ bytes.  In this
setting, many common approaches fail, simply because they cannot load
the dataset into memory or they are not sufficiently efficient. There
are roughly two approaches which can work:
\begin{enumerate}
\item Parallelize a batch learning algorithm over many machines (\textit{e.g.}, \cite{Chu08Map}).
\item Stream the examples to an online learning algorithm (\textit{e.g.},
    \cite{Littlestone88Learning}, \cite{Littlestone95Online},
    \cite{Cesabianchi96Worst}, and \cite{Kivinen97Exponentiated}).
\end{enumerate}
This paper focuses on the second approach.

Typical online learning algorithms have at least one weight for every
feature, which is too much in some applications for a couple reasons:

\begin{enumerate}
\item Space constraints. If the state of the online learning
algorithm overflows RAM it can not efficiently run.  A similar
problem occurs if the state overflows the L2 cache.
\item Test time constraints on computation. Substantially reducing
the number of features can yield substantial improvements in the
computational time required to evaluate a new sample.

\end{enumerate}
This paper addresses the problem of inducing sparsity in learned weights
while using an online learning algorithm. There are several ways to
do this wrong for our problem. For example:

\begin{enumerate}
\item Simply adding $L_{1}$ regularization to the gradient of an
online weight update doesn't work because gradients don't induce
sparsity.  The essential difficulty is that a gradient update has the
form $a+b$ where $a$ and $b$ are two floats.  Very few float pairs add
to $0$ (or any other default value) so there is little reason to
expect a gradient update to accidentally produce sparsity.
\item Simply rounding weights to $0$ is problematic because a weight may
be small due to being useless or small because it has been updated
only once (either at the beginning of training or because the set
of features appearing is also sparse). Rounding techniques can also
play havoc with standard online learning guarantees.
\item Black-box wrapper approaches which eliminate features and test
the impact of the elimination are not efficient enough.  These
approaches typically run an algorithm many times which is particularly
undesirable with large datasets.
\end{enumerate}

\subsection{What Others Do}

The Lasso algorithm \cite{Lasso} is commonly used to achieve $L_{1}$
regularization for linear regression. This algorithm does not work
automatically in an online fashion.
There are two formulations of $L_1$ regularization.
Consider a loss function $L(w,z_i)$ which is convex in $w$,
where $z_i=(x_i,y_i)$ is an input/output pair.
One is the convex constraint formulation
\begin{eqnarray}
\hat{w}=\arg\min_w \sum_{i=1}^n L(w, z_i) \quad \text{ subject to } \|w\|_1 \leq s , \label{eqn:lasso-constr}
\end{eqnarray}
where $s$ is a tunable parameter. The other is
soft-regularization, where
\begin{equation}
\hat{w}=\arg\min_w \sum_{i=1}^n L(w, z_i) + \lambda \|w\|_1 . \label{eqn:lasso}
\end{equation}
With appropriately chosen $\lambda$, the two formulations are equivalent.
The convex constraint formulation has a simple online version using the
projection idea in \cite{Zinkevich03}. It requires the projection of weight
$w$ into an $L_1$ ball at every online step.
This operation is difficult to implement efficiently
for large-scale data where we have examples with sparse features
and a large number of features. 
In such situation, we require that the number of operations per online step to be linear with respect to the number of nonzero features, and independent of 
the total number of features. Our method, which works with the
soft-regularization formulation (\ref{eqn:lasso}), satisfies the requirement.
Additional details can be found in Section~\ref{sec:implementation}. In addition to $L_1$ regularization formulation (\ref{eqn:lasso}), 
the family of online algorithms we consider
in this paper also include some nonconvex sparsification techniques.

The Forgetron algorithm \cite{Forgetron} is an online learning algorithm
that manages memory use. It operates by decaying the weights on previous
examples and then rounding these weights to zero when they become
small. The Forgetron is stated for kernelized online algorithms, while
we are concerned with the simple linear setting. When applied to a
linear kernel, the Forgetron is not computationally or space competitive
with approaches operating directly on feature weights.

\subsection{What We Do}

At a high level, the approach we take is weight decay to a default
value. This simple approach enjoys a strong performance guarantee, as
discussed in section \ref{sec:Theory}.  For instance, the algorithm
never performs much worse than a standard online learning algorithm,
and the additional loss due to sparsification is controlled
continuously with a single real-valued parameter.  The theory gives a
family of algorithms with convex loss functions for inducing
sparsity---one per online learning algorithm.  We instantiate this for
square loss and show how to deal with sparse examples efficiently in
section \ref{sec:Algorithm}.

As mentioned in the introduction, we are mainly interested in sparse
online methods for large scale problems with sparse features. 
For such problems, our algorithm should satisfy the following requirements:
\begin{itemize}
\item The algorithm should be computationally efficient: the number of operations per online step should be linear in the number of nonzero features, and independent of the total number of features.
\item The algorithm should be memory efficient: it needs to maintain a list of active features, and can insert (when the corresponding weight becomes nonzero) and delete (when the corresponding weight becomes zero) features dynamically.
\end{itemize}
The implementation details, showing that our methods satisfy the above requirements, are provided in section \ref{sec:implementation}.

Theoretical results stating how much sparsity is achieved using this
method generally require additional assumptions which may or may not
be met in practice.  Consequently, we rely on experiments in section
\ref{sec:Application} to show that our method achieves good sparsity
practice.  We compare our approach to a few others, including $L_1$
regularization on small data, as well as online rounding of
coefficients to zero.

\section{Online Learning with GD}

In the setting of standard online learning,
we are interested in sequential prediction problems where repeatedly
from $i=1,2,\ldots$:
\begin{enumerate}
\item An unlabeled example $x_{i}$ arrives.
\item We make a prediction based on existing weights $w_{i}\in R^{d}$.
\item We observe $y_i$, let $z_i=(x_i,y_i)$,
and incur some known loss $L(w_{i},z_{i})$ convex in parameter
$w_{i}$.
\item We update weights according to some rule: $w_{i+1}\leftarrow f(w_{i})$.
\end{enumerate}
We want to come up with an update rule $f$, which allows us to bound
the sum of losses\[\sum_{i=1}^{t}L(w_{i},z_{i})\] as well as achieving
sparsity. For this purpose, we start with the standard stochastic
gradient descent rule, which is of the form:
\begin{eqnarray}
f(w_i)= w_i - \eta \nabla_1 L (w_i,z_i) , \label{eq:online-update}
\end{eqnarray}
where $\nabla_1 L (a,b)$ is a sub-gradient of $L(a,b)$ with respect
to the first variable $a$. The parameter $\eta>0$ is often referred to
as the learning rate. In our analysis, we only consider constant learning
rate with fixed $\eta>0$ for simplicity. In theory, it might be
desirable to have a decaying learning rate $\eta_i$ which becomes smaller
when $i$ increases to get the so called {\em no-regret bound} without
knowing $T$ in advance. However, if $T$ is known in advance, one can
select a constant $\eta$ accordingly so that the regret vanishes as
$T \to \infty$. Since our focus is on
sparsity, not how to choose learning rate, for clarity, we use a
constant learning rate in the analysis because it leads to simpler bounds.

The above method has been widely used in online learning such as
\cite{Littlestone95Online} and \cite{Cesabianchi96Worst}.
Moreover, it is argued to be efficient even for solving batch problems
where we repeatedly run the online algorithm over training data
multiple times.  For example, the idea has been successfully applied to
solve large-scale standard SVM formulations~\cite{ShSiSr07,Zhang04-icml}.  In the scenario outlined
in the introduction, online learning methods are more suitable than
some traditional batch learning methods.

However, a main drawback of (\ref{eq:online-update}) is that it does
not achieve sparsity, which we address in this paper.
Note that in the literature, this particular update rule is often
referred to as gradient descent (GD) or stochastic gradient descent (SGD).
There are other variants, such as exponentiated gradient descent (EG).
Since our focus in this paper is sparsity, not GD versus EG, we shall
only consider modifications of (\ref{eq:online-update})
for simplicity.

\section{Sparse Online Learning}

\label{sec:Theory}

In this section, we examine several methods for achieving sparsity in
online learning. The first idea is simple coefficient rounding, which is the most
natural method. We will then consider its full online implementation,
and another method which is the online counterpart of $L_1$ regularization
in batch learning. As we shall see, all these ideas are closely related.

\subsection{Simple Coefficient Rounding} \label{sec:rounding}

In order to achieve sparsity, the most natural method is to round small
coefficients (that are no larger than a threshold $\theta>0$)
to zero after every $K$ online steps.
That is, if $i/K$ is not an integer, we use the standard GD rule in
(\ref{eq:online-update}); if $i/K$ is an integer, we modify the rule as:
\begin{equation}
f(w_i)= T_0(w_i - \eta \nabla_1 L (w_i,z_i), \theta) ,
\label{eq:sparse-online-update-rounding}
\end{equation}
where for a vector $v=[v_1,\ldots,v_d] \in R^d$, and a scalar
$\theta \geq 0$, $T_0(v,\theta)=[T_0(v_1,\theta),\ldots,T_0(v_d,\theta)]$, with
\[
T_0(v_j,\theta) =
\begin{cases}
0 & \text{if $|v_j| \leq \theta$} \\
v_j & \text{otherwise}
\end{cases} .
\]
That is, we first perform a standard stochastic gradient descent rule,
and then round the updated coefficients toward zero.
The effect is to remove nonzero and small components in the weight vector.

In general, we should not take $K=1$, especially when $\eta$ is small,
since each step modifies $w_i$ by only a small amount.  If a
coefficient is zero, it remains small after one online update, and the
rounding operation pulls it back to zero.  Consequently, rounding can
be done only after every $K$ steps (with a reasonably large $K$); in
this case, nonzero coefficients have sufficient time to go above the
threshold $\theta$.  However, if $K$ is too large, then in the
training stage, we will need to keep many more nonzero features in the
intermediate steps before they are rounded to zero. In the extreme
case, we may simply round the coefficients in the end, which does not
solve the storage problem in the training phase.  The sensitivity in
choosing appropriate $K$ is a main drawback of this method; another
drawback is the lack of theoretical guarantee for its online
performance.

The above mentioned issues motivate us to consider more principled
sparse online learning methods.  In section~\ref{sec:truncatedgradient},
we derive an online version of rounding using an idea
called \emph{truncated gradient} for which regret bounds hold.

\subsection{A Sub-gradient Algorithm for $L_1$ Regularization}
\label{sec:subgradient}

In our experiments, we combine rounding-in-the-end-of-training with a simple
online sub-gradient method for $L_1$ regularization with a
regularization parameter $g>0$:
\begin{equation}
f(w_i)= w_i - \eta \nabla_1 L (w_i,z_i) - \eta g \, \sgn(w_i) , \label{eq:online-subgradient-L1}
\end{equation}
where for a vector $v=[v_1,\ldots,v_d]$, $\sgn(v)=[\sgn(v_1),\ldots,\sgn(v_d)]$,
and $\sgn(v_j)=1$ when $v_j>0$, $\sgn(v_j)=-1$ when $v_j<0$, and
$\sgn(v_j)=0$ when $v_j=0$.
In the experiments, the online
method (\ref{eq:online-subgradient-L1}) plus rounding in the end
is used as a simple baseline. One should note that
this method does not produce sparse weights online. Therefore it does not
handle
large-scale problems for which we cannot keep all features in memory.

\subsection{Truncated Gradient} \label{sec:truncatedgradient}

In order to obtain an online version of the simple rounding rule in
(\ref{eq:sparse-online-update-rounding}), we observe that the direct
rounding to zero is too aggressive. A less aggressive version is
to shrink the coefficient to zero by a smaller amount. We call this idea
truncated gradient.

The amount of shrinkage is measured by a \emph{gravity} parameter
$g_i>0$:
\begin{equation}
f(w_i)= T_1(w_i - \eta \nabla_1 L (w_i,z_i), \eta g_i, \theta) ,
\label{eq:sparse-online-update-truncated-grad}
\end{equation}
where for a vector $v=[v_1,\ldots,v_d] \in R^d$, and a scalar
$g \geq 0$, $T_1(v,\alpha,\theta)=[T_1(v_1,\alpha,\theta),\ldots,T_1(v_d,\alpha,\theta)]$, with
\[
T_1(v_j,\alpha,\theta) =
\begin{cases}
\max(0,v_j- \alpha) & \text{if $v_j \in [0,\theta]$} \\
\min(0,v_j + \alpha) & \text{if $v_j \in [-\theta,0]$} \\
v_j & \text{otherwise}
\end{cases} .
\]
Again, the truncation can be performed every $K$ online steps.
That is, if $i/K$ is not an integer, we let $g_i=0$;
if $i/K$ is an integer, we let $g_i=Kg$ for a gravity parameter
$g>0$.
This particular choice is equivalent to (\ref{eq:sparse-online-update-rounding})
when we set $g$ such that $\eta K g \geq \theta$.
This requires a large $g$ when $\eta$ is small.
In practice, one should set a small, fixed $g$, as implied by our
regret bound developed later.

In general, the larger the parameters $g$ and $\theta$ are, the more
sparsity is incurred.  Due to the extra truncation $T_1$, this method
can lead to sparse solutions, which is confirmed in our experiments
described later.  In those experiments, the degree of sparsity
discovered varies with the problem.

A special case, which we use in the experiment, is to let $g=\theta$
in (\ref{eq:sparse-online-update-truncated-grad}). In this case,
we can use only one parameter $g$ to control sparsity.
Since $\eta K g \ll \theta$ when $\eta K$ is small,
 the truncation operation is less aggressive than the rounding
in (\ref{eq:sparse-online-update-rounding}).
At first sight, the procedure appears to be an ad-hoc way to fix
(\ref{eq:sparse-online-update-rounding}).
However, we can establish a regret bound for this method,
showing that it is theoretically sound.

Another important special case of (\ref{eq:sparse-online-update-truncated-grad})
is setting $\theta=\infty$.
This leads to the following update rule for every $K$-th online step
\begin{eqnarray}
f(w_i)= T(w_i - \eta \nabla_1 L (w_i,z_i), g_i \eta) , \label{eq:sparse-online-update}
\end{eqnarray}
where for a vector $v=[v_1,\ldots,v_d] \in R^d$, and a scalar
$g \geq 0$, $T(v,\alpha)=[T(v_1,\alpha),\ldots,T(v_d, \alpha)]$, with
\[
T(v_j,\alpha) =
\begin{cases}
\max(0,v_j - \alpha) & \text{if $v_j>0$} \\
\min(0,v_j + \alpha) & \text{otherwise}
\end{cases} .
\]
The method is a modification of the standard
sub-gradient online method for $L_1$ regularization in
(\ref{eq:online-subgradient-L1}).
The parameter $g_i \geq 0$ controls the sparsity that can be achieved
with the algorithm.  Note that when $g_i=0$, the update rule is
identical to the standard stochastic gradient descent rule.
In general, we may perform a truncation every $K$ steps.
That is, if $i/K$ is not an integer, we let $g_i=0$;
if $i/K$ is an integer, we let $g_i=Kg$ for a gravity parameter
$g>0$. The reason for doing so (instead of a constant $g$) is that
we can perform a more aggressive truncation with gravity parameter
$Kg$ after each $K$ steps. This can potentially lead to better sparsity.

The procedure in (\ref{eq:sparse-online-update})
can be regarded as an online counterpart of $L_1$ regularization in the
sense that it approximately solves an $L_1$ regularization problem in
the limit of $\eta \to 0$.  Truncated gradient descent for $L_1$
regularization is different from the naive application of stochastic
gradient descent rule (\ref{eq:online-update}) with an added $L_1$
regularization term. As pointed out in the introduction, the latter
fails because it rarely leads to sparsity.  Our theory shows that
even with sparsification, the prediction performance is still
comparable to that of the standard online learning algorithm.  In the
following, we develop a general regret bound for this general method,
which also shows how the regret may depend on the sparsification
parameter $g$.

\subsection{Regret Analysis}

Throughout the paper, we use $\|\cdot\|_1$ for $1$-norm, and
$\|\cdot\|$ for $2$-norm.  For reference, we make the
following assumption regarding the loss function:
\begin{assumption}
  We assume that $L(w,z)$ is convex in $w$, and
  there exist non-negative constants $A$ and $B$ such that
  $(\nabla_1 L(w,z))^2 \leq A L(w,z) + B$ for all $w \in R^d$ and $z \in R^{d+1}$.
  \label{assump:loss}
\end{assumption}
For linear prediction problems, we have a general loss function
of the form $L(w,z)= \phi(w^T x, y)$.
The following are some common loss functions
$\phi(\cdot,\cdot)$ with corresponding choices of
parameters $A$ and $B$ (which are not unique), under the assumption that
$\sup_x \|x\| \leq C$.
\begin{itemize}
\item Logistic: $\phi(p,y)=\ln (1+\exp(-py))$; $A=0$ and $B=C^2$.
  This loss is for binary classification problems with $y \in \{\pm 1\}$.
\item SVM (hinge loss):  $\phi(p,y)=\max(0,1-py)$; $A=0$ and $B=C^2$.
  This loss is for binary classification problems with $y \in \{\pm 1\}$.
\item Least squares (square loss): $\phi(p,y)=(p-y)^2$; $A=4 C^2$ and $B=0$.
  This loss is for regression problems.
\end{itemize}

Our main result is Theorem \ref{thm:sparse-online-regret} that
is parameterized by $A$ and $B$.  The proof is left to the appendix.
Specializing it to particular losses yields several
corollaries.  The one that can be applied to the least squares loss
will be given later in Corollary~\ref{cor:sparse-squared}.

\begin{thm} (Sparse Online Regret) \label{thm:sparse-online-regret}
Consider sparse online update rule (\ref{eq:sparse-online-update})
with $w_1=0$ and $\eta>0$.  If Assumption~\ref{assump:loss} holds,
then for all $\bar{w} \in R^d$
we have
\begin{align*}
&\frac{1-0.5 A\eta}{T}\sum_{i=1}^T \left[L(w_i,z_i) + \frac{g_i}{1-0.5 A\eta} \|w_{i+1} \cdot I(w_{i+1} \leq \theta) \|_1 \right] \\
\leq& \frac{\eta}{2} B + \frac{\|\bar{w}\|^2}{2\eta T} + \frac{1}{T}\sum_{i=1}^T [L(\bar{w},z_i) + g_i \| \bar{w} \cdot I(w_{i+1} \leq \theta) \|_1],
\end{align*}
where for  vectors $v=[v_1,\ldots,v_d]$ and $v'=[v_1',\ldots,v_d']$, we let
\[
\| v \cdot I(|v'| \leq \theta)\|_1 = \sum_{j=1}^d |v_j| I(|v_j'| \leq \theta),
\]
where $I(\cdot)$ is the set indicator function.
\end{thm}

We state the theorem with a constant learning rate $\eta$. As mentioned
earlier, it is possible to obtain a result with variable learning rate
where $\eta=\eta_i$ decays as $i$ increases. Although this may lead
to a no-regret bound without knowing $T$ in advance,
it introduces extra complexity to the presentation of the main idea.
Since our focus is on sparsity rather than optimizing learning rate, we do not include
such a result for clarity. If $T$ is known in advance, then in the
above bound, one can simply take $\eta=O(1/\sqrt{T})$ and the regret is of
order $O(1/\sqrt{T})$.

In the above theorem, the right-hand side involves a term $g_i \|
\bar{w} \cdot I(w_{i+1} \leq \theta) \|_1$ that depends on $w_{i+1}$
which is not easily estimated.  To remove this dependency, a trivial
upper bound of $\theta=\infty$ can be used, leading to $L_1$ penalty
$g_i \|\bar{w}\|_1$.  In the general case of $\theta<\infty$, we
cannot remove the $w_{i+1}$ dependency because the effective
regularization condition (as shown on the left-hand side) is the
non-convex penalty $g_i \|w \cdot I(|w| \leq \theta)\|_1$.  Solving
such a non-convex formulation is hard both in the online and batch
settings.  In general, we only know how to efficiently discover a
local minimum which is difficult to characterize.  Without a good
characterization of the local minimum, it is not possible for us to
replace $g_i \| \bar{w} \cdot I(w_{i+1} \leq \theta) \|_1$ on the
right-hand side by $g_i \| \bar{w} \cdot I(\bar{w} \leq \theta) \|_1$
because such a formulation would have implied that we could
efficiently solve a non-convex problem with a simple online update
rule.  Still, when $\theta <\infty$, one naturally expects that the
right-hand side penalty $g_i \| \bar{w} \cdot I(w_{i+1} \leq \theta)
\|_1$ is much smaller than the corresponding $L_1$ penalty $g_i
\|\bar{w}\|_1$, especially when $w_j$ has many components that are close to $0$.
Therefore the situation with $\theta<\infty$ can potentially
yield better performance on some data. This is confirmed in our experiments.

Theorem~\ref{thm:sparse-online-regret} also
implies a trade-off between sparsity
and regret performance. We may simply consider the case
where $g_i=g$ is a constant.
When $g$ is small, we have less sparsity but the
regret term $g \| \bar{w} \cdot I(w_{i+1} \leq \theta) \|_1 \leq
g\|\bar{w}\|_1$ on the right-hand side is also small.
When $g$ is large, we are able to achieve more sparsity but the regret
$ g\| \bar{w} \cdot I(w_{i+1} \leq \theta) \|_1$
on the right-hand side also becomes large.
Such a trade-off (sparsity versus prediction accuracy) is empirically studied in Section~\ref{sec:Application}.
Our observation suggests that we can gain significant sparsity with only a small decrease of accuracy (that is, using a small $g$).

Now consider the case $\theta =\infty$ and $g_i=g$.
When $T \to \infty$, if we let $\eta \to 0$ and $\eta T \to
\infty$, then Theorem~\ref{thm:sparse-online-regret} implies that
\begin{align*}
\frac{1}{T}\sum_{i=1}^T [L({w}_i,z_i) + g \|w_i\|_1]
\leq \inf_{\bar{w} \in R^d}
\left[\frac{1}{T}\sum_{i=1}^T L(\bar{w},z_i) + 2g \| \bar{w} \|_1 \right]  + o(1).
\end{align*}
In other words, if we let $L'(w,z)=L(w,z)+g\|w\|_1$ be the $L_1$ regularized
loss, then the $L_1$ regularized regret is small when $\eta \to 0$ and
$T \to \infty$. This implies that our procedure can be regarded as the
online counterpart of $L_1$-regularization methods.  In the stochastic
setting where the examples are drawn iid from some underlying
distribution, the sparse online gradient method proposed in this paper
solves the $L_1$ regularization problem.

\subsection{Stochastic Setting}

Stochastic-gradient-based online learning methods can be used to solve
large-scale batch optimization problems, often quite successfully
\cite{ShSiSr07,Zhang04-icml}.  In this setting, we can go through
training examples one-by-one in an online fashion, and repeat multiple
times over the training data. In this section, we analyze the
performance of such a procedure using
Theorem~\ref{thm:sparse-online-regret}.

To simplify the analysis,
instead of assuming that we go through the data one by one, we assume that
each additional data point is drawn from the training data randomly with equal probability. This corresponds to the standard stochastic optimization setting, in which observed samples are iid from some underlying distributions.
The following result is
a simple consequence of Theorem~\ref{thm:sparse-online-regret}.
For simplicity, we only consider the case with $\theta=\infty$
and constant gravity $g_i=g$.
\begin{thm} \label{thm:sparse-online-stochastic}
  Consider a set of training data $z_i=(x_i,y_i)$ for $i=1,\ldots,n$, and let
  \[
  R(w,g) = \frac{1}{n}\sum_{i=1}^n L(w,z_i) + g \|w\|_1
  \]
  be the $L_1$ regularized loss over training data.
  Let $\hat{w}_1=w_1=0$, and define recursively for $t = 1, 2,\ldots$
  \begin{align*}
  w_{t+1} =T(w_{t} - \eta \nabla_1(w_t,z_{i_t}), g \eta), \qquad
  \hat{w}_{t+1} = \hat{w}_{t} + (w_{t+1}-\hat{w}_{t})/(t+1) ,
\end{align*}
where each $i_t$ is drawn from $\{1,\ldots,n\}$ uniformly at random.
If  Assumption~\ref{assump:loss} holds,
then at any time $T$, the following inequalities are valid for all $\bar{w} \in R^d$:
\begin{align*}
&\E_{i_1,\ldots,i_T} \left[(1-0.5 A\eta) R\left(\hat{w}_T,\frac{g}{1-0.5 A\eta}\right)\right] \\
\leq &\E_{i_1,\ldots,i_T} \left[\frac{1-0.5 A\eta}{T}\sum_{i=1}^T R\left(w_i,\frac{g}{1-0.5 A\eta}\right)\right] \\
\leq &\frac{\eta}{2} B + \frac{\|\bar{w}\|^2}{2\eta T} + R(\bar{w},g) .
\end{align*}
\end{thm}
\begin{proof}
Note that the recursion of $\hat{w}_t$ implies that
\[
\hat{w}_T= \frac{1}{T} \sum_{t=1}^T w_t
\]
from telescoping the update rule.

Because $R(w,g)$ is convex in $w$, the first inequality follows
directly from Jensen's inequality.

In the following we only need to prove the second inequality.
Theorem~\ref{thm:sparse-online-regret} implies the following:
\begin{align}
\frac{1-0.5 A\eta}{T}\sum_{t=1}^T \left[L(w_t,z_{i_t}) + \frac{g}{1-0.5 A\eta} \|w_{t}\|_1 \right]
\leq g \|\bar{w}\|_1 + \frac{\eta}{2} B + \frac{\|\bar{w}\|^2}{2\eta T} + \frac{1}{T}\sum_{t=1}^T L(\bar{w},z_{i_t}) . \label{eq:stoch-one-instance}
\end{align}
Observe that
\begin{align*}
 \E_{i_t} \left[L(w_t,z_{i_t}) + \frac{g}{1-0.5 A\eta} \|w_{t}\|_1 \right]
= R\left(w_t,\frac{g}{1-0.5A\eta}\right)
\end{align*}
and
\[
g \|\bar{w}\|_1 + \E_{i_1,\ldots,i_T}\left[\frac{1}{T}\sum_{t=1}^T L(\bar{w},z_{i_t})\right]
= R(\bar{w},g) .
\]
The second inequality is obtained by taking the expectation with respect to
$\E_{i_1,\ldots,i_T}$ in (\ref{eq:stoch-one-instance}).
\end{proof}

If we let $\eta \to 0$ and $\eta T \to \infty$, the bound in
Theorem~\ref{thm:sparse-online-stochastic} becomes
\[
\E \left[R(\hat{w}_T,g)\right] \leq \E \left[\frac{1}{T}\sum_{t=1}^T R(w_t,g)\right] \leq \inf_{\bar{w}} R(\bar{w},g) + o(1) .
\]
That is, on average $\hat{w}_T$ approximately solves the $L_1$ regularization problem
\[
\inf_w \left[\frac{1}{n}\sum_{i=1}^n L(w,z_i) + g \|w\|_1 \right] .
\]
If we choose a random stopping time $T$, then the above inequalities
says that on average $R(w_T)$ also solves this $L_1$ regularization
problem approximately.  Therefore in our experiment, we use the last
solution $w_T$ instead of the aggregated solution $\hat{w}_T$.

Since 1-norm regularization is frequently used to achieve sparsity in the batch learning setting, the connection to 1-norm regularization
can be regarded as an alternative
justification for the sparse-online algorithm developed in this paper.

\section{Truncated Gradient Algorithm for Least Squares} \label{sec:Algorithm}

The method in Section~\ref{sec:Theory} can be directly applied
to least squares regression. This leads to
Algorithm \ref{alg:SLS} which implements sparsification for square loss according to equation (\ref{eq:sparse-online-update}).
In the description, we use superscripted symbol $w^j$ to denote the $j$-th
component of vector $w$ (in order to differentiate from $w_i$, which we
have used to denote the $i$-th weight vector).
For clarity, we also drop the index $i$ from $w_i$.
Although we keep the choice of gravity parameters $g_i$ open
in the algorithm description, in practice, we only consider the
following choice:
\[
g_i = \begin{cases}
K g & \text{ if $i/K$ is an integer} \\
0   & \text{ otherwise}
\end{cases} .
\]
This may give a more aggressive truncation (thus sparsity) after every $K$-th
iteration. Since we do not have a theorem formalizing how much more
sparsity one can gain from this idea, its effect will only be examined
through experiments.

\begin{algorithm*}[t]
\caption{Truncated Gradient} \label{alg:SLS}
{\bf Inputs:}
\begin{itemize}
\item threshold $\theta \ge 0$
\item gravity sequence $g_i \ge 0$
\item learning rate $\eta \in (0,1)$
\item example oracle $\mathcal{O}$
\end{itemize}

{\bf initialize} weights $w^j\leftarrow 0$ ($j=1,\ldots,d$)

{\bf for} trial $i=1,2,\ldots$

\begin{enumerate}
\item Acquire an unlabeled example $x=[x^1,x^2,\ldots,x^d]$ from oracle $\mathcal{O}$
\item {\bf forall} weights $w^j$ ($j=1,\ldots,d$)
\begin{enumerate}
\item {\bf if} $w^j>0$ and $w^j \leq \theta$
{\bf then} $w^{j}\leftarrow\max\{ w^{j}-g_i\eta,0\}$
\item {\bf elseif} $w^j < 0$ and $w^j \geq -\theta$
{\bf then} $w^{j}\leftarrow\min\{ w^{j}+g_i\eta,0\}$
\end{enumerate}
\item Compute prediction: $\hat{y}=\sum_{j}w^{j}x^{j}$
\item Acquire the label $y$ from oracle $\mathcal{O}$
\item Update weights for all features $j$: $w^{j}\leftarrow w^{j}+2\eta(y-\hat{y})x^{j}$
\end{enumerate}
\end{algorithm*}

In many online
learning situations (such as web applications), only a small subset of
the features have nonzero values for any example $x$.  It is thus
desirable to deal with sparsity only in this small subset rather than
all features, while simultaneously inducing sparsity on all feature
weights.  Moreover, it is important to store only features with
non-zero coefficients (if the number of features is so large that
it cannot be stored in memory, this approach allows us to use a hashtable
to track only the nonzero coefficients).
We describe how this can be implemented efficiently in the next section.

For reference, we present a specialization of
Theorem \ref{thm:sparse-online-regret}
in the following corollary that is directly applicable to
 Algorithm \ref{alg:SLS}.
\begin{cor}
\label{cor:sparse-squared} (Sparse Online Square Loss Regret)
If there exists $C>0$ such that for all $x$, $\|x\| \leq C$,
then for all $\bar{w} \in R^d$, we have
\begin{align*}
&\frac{1- 2 C^2\eta}{T}\sum_{i=1}^T
\left[(w_i^T x_i -y_i)^2  + \frac{g_{i}}{1-2 C^2 \eta} \|w_i \cdot I(|w_i| \leq \theta) \|_1 \right] \\
\leq&  \frac{\|\bar{w}\|^2}{2 \eta T} + \frac{1}{T}\sum_{i=1}^T
\left[ (\bar{w}^T x_i-y_i)^2 + g_{i+1} \|\bar{w}\cdot I(|w_{i+1}| \leq \theta) \|_1 \right] ,
\end{align*}
where $w_i=[w^1,\ldots,w^d] \in R^d$ is the weight vector used for prediction
at the $i$-th step of Algorithm~\ref{alg:SLS};
$(x_i,y_i)$ is the data point observed at the $i$-step.
\end{cor}
This corollary explicitly states that the average square loss
incurred by the learner (left term) is bounded by the average square loss of the best weight vector $\bar{w}$, plus a term related to the
size of $\bar{w}$ which decays as $1/T$ and an additive offset
controlled by the sparsity threshold $\theta$ and the
gravity parameter $g_i$.

\section{Efficient Implementation}
\label{sec:implementation}

We altered a standard gradient-descent implementation (\vw~\citep{VW})
according to algorithm \ref{alg:SLS}.  \vw optimizes square loss on a
linear representation $w \cdot x$ via gradient descent (\ref{eq:online-update})
with a couple caveats:
\begin{enumerate}
\item The prediction is normalized by the square root of the number of
  nonzero entries in a sparse vector, $w \cdot x / |x|_0^{0.5}$.  This
  alteration is just a constant rescaling on dense vectors which is
  effectively removable by an appropriate rescaling of the learning
  rate.
\item The prediction is clipped to the interval $[0,1]$, implying that
  the loss function is not square loss for unclipped predictions
  outside of this dynamic range.  Instead the update is a constant
  value, equivalent to the gradient of a linear loss function.
\end{enumerate}
The learning rate in \vw is controllable, supporting $1/i$ decay as
well as a constant learning rate (and rates in-between).  The program
operates in an entirely online fashion, so the memory footprint is
essentially just the weight vector, even when the amount of data is
very large.

As mentioned earlier, we would like the algorithm's computational
complexity to depend linearly on the number of nonzero features of an
example, rather than the total number of features.
The approach we took was to store a time-stamp $\tau_j$ for
each feature $j$.  The time-stamp was initialized to the index
of the example where feature $j$ was nonzero for the first time.
During online learning, we simply went through all nonzero features
$j$ of example $i$, and could ``simulate'' the shrinkage of $w^j$
after $\tau_j$ in a batch mode.
These weights are then updated, and
their time stamps are reset to $i$.  This lazy-update idea of
delaying the shrinkage calculation until needed is the key to
efficient implementation of truncated gradient.
Specifically, instead of using update rule
(\ref{eq:sparse-online-update-truncated-grad}) for weight $w^j$,
we shrunk the weights of all nonzero feature $j$ differently by
the following:
\[
f(w^j) = T_1\left(w^j+2\eta(y-\hat{y})x^j,\left\lfloor\frac{i-\tau_j}{K}\right\rfloor K \eta g,\theta\right),
\]
and $\tau_j$ is updated by
\[
\tau_j \leftarrow \tau_j + \left\lfloor\frac{i-\tau_j}{K}\right\rfloor K.
\]

We note that such a lazy-update trick by maintaining the
time-stamp information can be applied to the other two
algorithms given in section~\ref{sec:Theory}.  In the
coefficient rounding algorithm (\ref{eq:sparse-online-update-rounding}),
for instance, for each nonzero feature $j$ of example $i$, we can first
perform a regular gradient descent on the square loss,
and then do the following: if $|w_j|$ is below the threshold
$\theta$ and $i\ge\tau_j+K$, we round $w_j$ to $0$ and set
$\tau_j$ to $i$.

This implementation shows that the truncated gradient method satisfies
the following requirements needed for solving large scale problems with
sparse features.
\begin{itemize}
\item The algorithm is computationally efficient: the number of operations per online step is linear in the number of nonzero features, and independent of the total number of features.
\item The algorithm is memory efficient: it maintains a list of active features, and a feature can be inserted when observed, and deleted when the corresponding weight becomes zero.
\end{itemize}

If we apply the online projection idea in \cite{Zinkevich03}
to solve (\ref{eqn:lasso-constr}), then in the update rule (\ref{eq:sparse-online-update}), one has to pick the smallest $g_i \geq 0$
such that $\|w_{i+1}\|_1 \leq s$. We do not know an efficient method to find this specific $g_i$ using operations independent of the total number of features. A standard implementation relies on sorting all weights, which requires $O(d \ln d)$ operations, where $d$ is the total number of (nonzero) features. This complexity is unacceptable for our purpose. However, we shall point out that in an important recent work \cite{DuShSiCh08}, the authors proposed an efficient online $\ell_1$-projection method.
The idea is to use a balanced tree to keep track of weights, which allows efficient threshold finding and tree updates in $O(k \ln d)$ operations on average (here, $k$ denotes the number of nonzero coefficients in the current training example). Although the algorithm still has weak dependency on $d$, it is applicable to large scale practical applications. 
The theoretical analysis presented in this paper shows that we can obtain a meaningful regret bound by picking an arbitrary $g_i$. This is useful because the resulting method is much simpler to implement and is computationally more efficient per online step. Moreover, our method allows non-convex updates that are closely related to the simple coefficient rounding idea. Due to the complexity of implementing the balanced tree strategy in \cite{DuShSiCh08}, we shall not compare to it in this paper. 

\section{Empirical Results} \label{sec:Application}

We applied \vw with the efficiently implemented sparsify option, as described in the
previous section, to a selection of datasets, including eleven
datasets from the UCI repository~\citep{Asuncion07Uci}, the
much larger dataset rcv1~\citep{Lewis04Rcv1}, and a private
large-scale dataset \bigads related to ad interest prediction.
While UCI datasets are useful for benchmark purposes, rcv1 and
\bigads are more interesting since they embody real-world datasets
with large numbers of features, many of which are less informative
for making predictions than others.  The datasets are summarized
in Table~\ref{tbl:datasets}.

The UCI datasets we used do not have many features, and it is
expected that a large fraction of these features are useful for
making predictions.  For comparison purposes as well as to better
demonstrate the behavior of our algorithm, we also added $1000$
random binary features to those datasets.  Each feature has value
$1$ with probability $0.05$ and $0$ otherwise.

\begin{table}[t]
\begin{center}
\caption{Dataset Summary.} \vspace{0.1in} \label{tbl:datasets}
\begin{tabular}{|c|cccc|}
\hline Dataset& \#features& \#train data& \#test data& task
\tabularnewline \hline \hline ad& 1411& 2455& 824&
classification\tabularnewline \hline crx& 47& 526& 164&
classification\tabularnewline \hline housing& 14& 381& 125&
regression\tabularnewline \hline krvskp& 74& 2413& 783&
classification\tabularnewline \hline magic04& 11& 14226& 4794&
classification\tabularnewline \hline mushroom& 117& 6079& 2045&
classification\tabularnewline \hline spambase& 58& 3445& 1156&
classification\tabularnewline \hline wbc& 10& 520& 179&
classification\tabularnewline \hline wdbc& 31& 421& 148&
classification\tabularnewline \hline wpbc& 33& 153& 45&
classification\tabularnewline \hline zoo& 17& 77& 24&
regression\tabularnewline \hline rcv1& 38853& 781265& 23149&
classification\tabularnewline \hline \bigads& $3 \times 10^9$& $26
\times 10^6$& $2.7 \times 10^6$& classification\tabularnewline
\hline
\end{tabular}
\end{center}
\end{table}

\subsection{Feature Sparsification of Truncated Gradient Descent}

In the first set of experiments, we are interested in how much
reduction in the number of features is possible without affecting
learning performance significantly; specifically, we require the
accuracy be reduced by no more than $1\%$ for classification
tasks, and the total square loss be increased by no more than
$1$\% for regression tasks.  As common practice, we allowed the
algorithm to run on the training data set for multiple passes with
decaying learning rate. For each dataset, we performed $10$-fold
cross validation over the training set to identify the best set of
parameters, including the learning rate $\eta$, the sparsification
rate $g$, number of passes of the training set, and the decay of
learning rate across these passes.  This set of parameters was
then used to train \vw on the whole training set.  Finally, the
learned classifier/regressor is evaluated on the test set.  We
fixed $K=1$ and $\theta=\infty$ in these experiments, and will
study the effects of $K$ and $\theta$ in later subsections.

\begin{figure}[t]
\begin{center}
\includegraphics[angle=270,width=0.55\columnwidth]{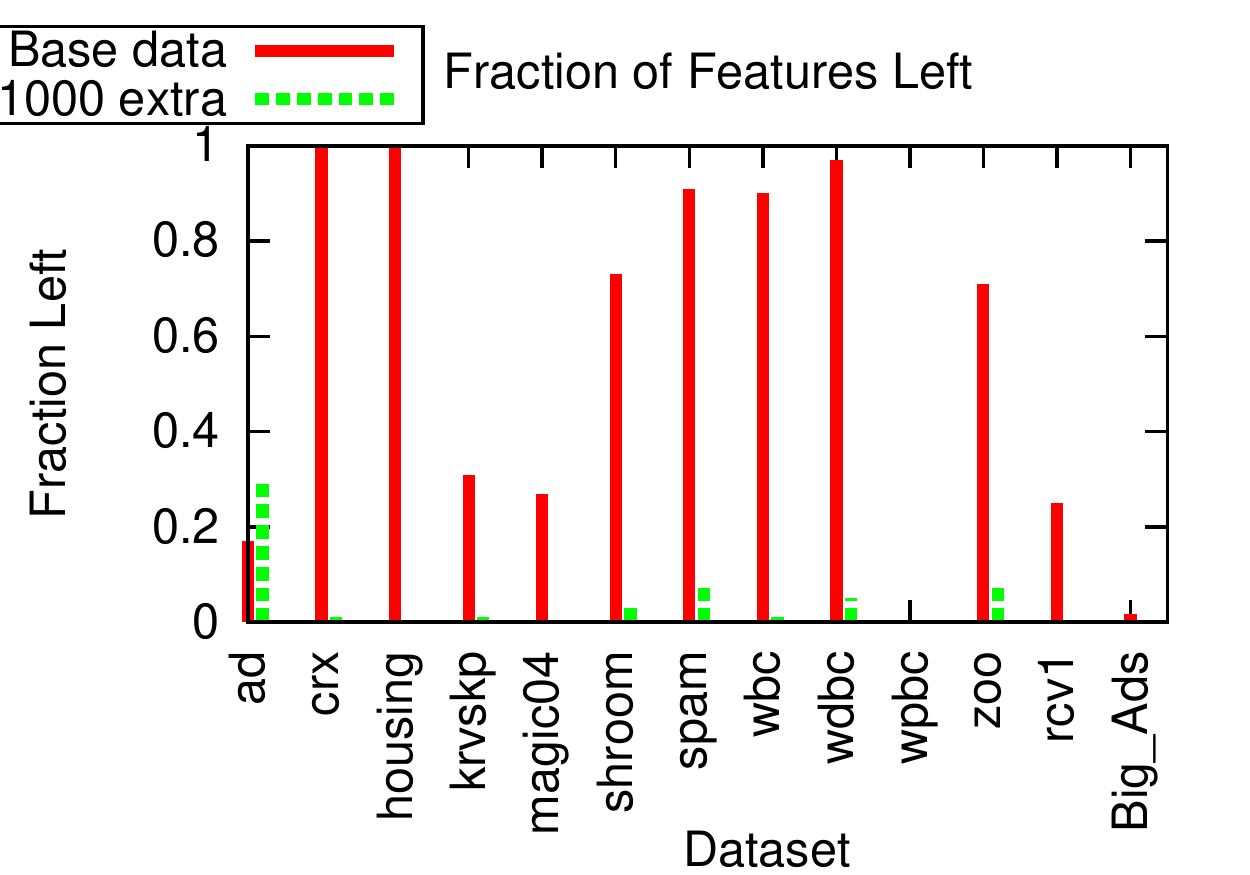}
\end{center}
\caption{\label{fig:Features} A plot showing the amount of
features left after sparsification for each dataset. The first
result is the fraction of features left when the performance is
changed by at most 1\% due to sparsification. The second result is
the \% sparsification when $1000$ random features are added to
each example.  For rcv1 and \bigads there is no second column,
since the experiment is not useful.}
\end{figure}

Figure~\ref{fig:Features} shows the fraction of reduced features
after sparsification is applied to each dataset.  For UCI datasets
with randomly added features, \vw is able to reduce the number of
features by a fraction of more than $90\%$, except for the ad
dataset in which only $71\%$ reduction is observed.  This less
satisfying result might be improved by a more extensive parameter
search in cross validation. However, if we can tolerate $1.3\%$
decrease in accuracy (instead of $1\%$ as for other datasets)
during cross validation, \vw is able to achieve $91.4\%$
reduction, indicating that a large reduction is still possible at
the tiny additional cost of $0.3\%$ accuracy loss. With this
slightly more aggressive sparsification, the test-set accuracy
drops from $95.9\%$ (when only $1\%$ loss in accuracy is allowed
in cross validation) to $95.4\%$, while the accuracy without
sparsification is $96.5\%$.

Even for the original UCI datasets without artificially added
features, \vw manages to filter out some of the less useful
features while maintaining the same level of performance.  For
example, for the ad dataset, a reduction of $83.4\%$ is achieved.
Compared to the results above, it seems the most effective feature
reductions occur on datasets with a large number of less useful
features, exactly where sparsification is needed.

For rcv1, more than $75\%$ of features are removed after the
sparsification process, indicating the effectiveness of our
algorithm in real-life problems.  We were not able to try many
parameters in cross validation because of the size of rcv1. It is
expected that more reduction is possible when a more thorough
parameter search is performed.

The previous results do not exercise the full power of the
approach presented here because they are applied to datasets where
standard Lasso regularization \citep{Lasso} is or may be
computationally viable. We have also applied this approach to a
large non-public dataset \bigads where the goal is predicting
which of two ads was clicked on given context information (the
content of ads and query information).  Here, accepting a $0.009$
increase in classification error allows us to reduce the number of
features from about $3\times10^9$ to about $24\times10^6$, a factor
of $125$ decrease in the number of features.

\begin{figure}[t]
\begin{center}
\includegraphics[angle=270,width=0.55\columnwidth]{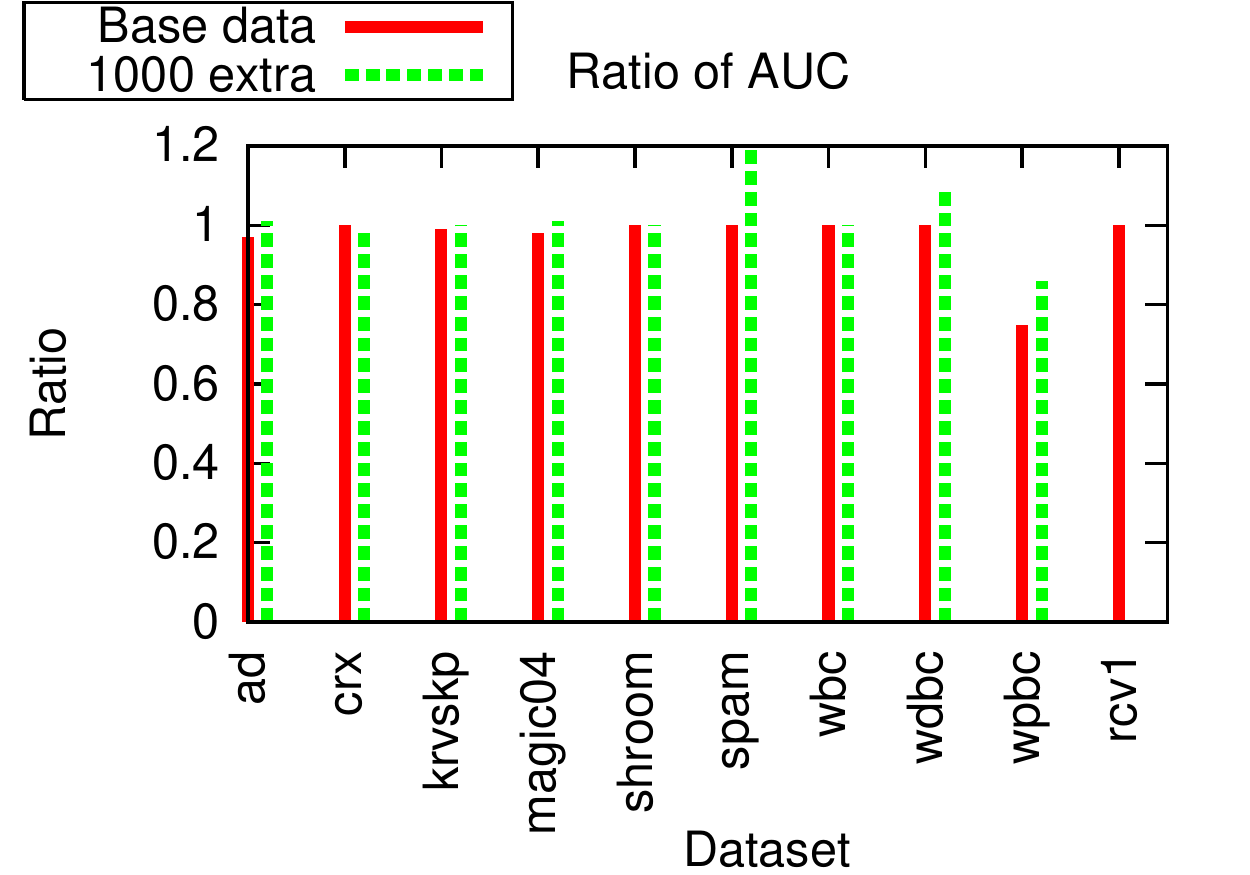}
\end{center}
\caption{\label{fig:auc} A plot showing the ratio of the AUC when
sparsification is used over the AUC when no sparsification is
used. The same process as in Figure~\ref{fig:Features} is used to
determine empirically good parameters.  The first result is for
the original dataset, while the second result is for the modified
dataset where $1000$ random features are added to each example. }
\end{figure}

For classification tasks, we also study how our sparsification
solution affects 
AUC (Area Under the ROC Curve), which is a standard metric 
for classification.\footnote{We use AUC here and in later
subsections because it is insensitive to threshold, which is unlike
accuracy. }
Using the same sets of
parameters from $10$-fold cross validation described above, we
find that the criterion is not affected significantly by
sparsification and in some cases, they are actually slightly
improved.  The reason may be that our sparsification method remove
some of the features that could have confused \vw.  The ratios of
the AUC with and without sparsification for all
classification tasks are plotted in Figures~\ref{fig:auc}.
It is often the case that these
ratios are above $98\%$.  

\subsection{The Effects of $K$}

As we argued before, using a $K$ value larger than $1$ may be
desired in truncated gradient and the rounding algorithms.  This
advantage is empirically demonstrated here.  In particular, we try
$K=1$, $K=10$, and $K=20$ in both algorithms.  As before, cross
validation is used to select parameters in the rounding algorithm,
including learning rate $\eta$, number of passes of data during
training, and learning rate decay over training passes.

\begin{figure}[t]
\begin{center}
\begin{tabular}{cc}
\includegraphics[width=0.4\columnwidth]{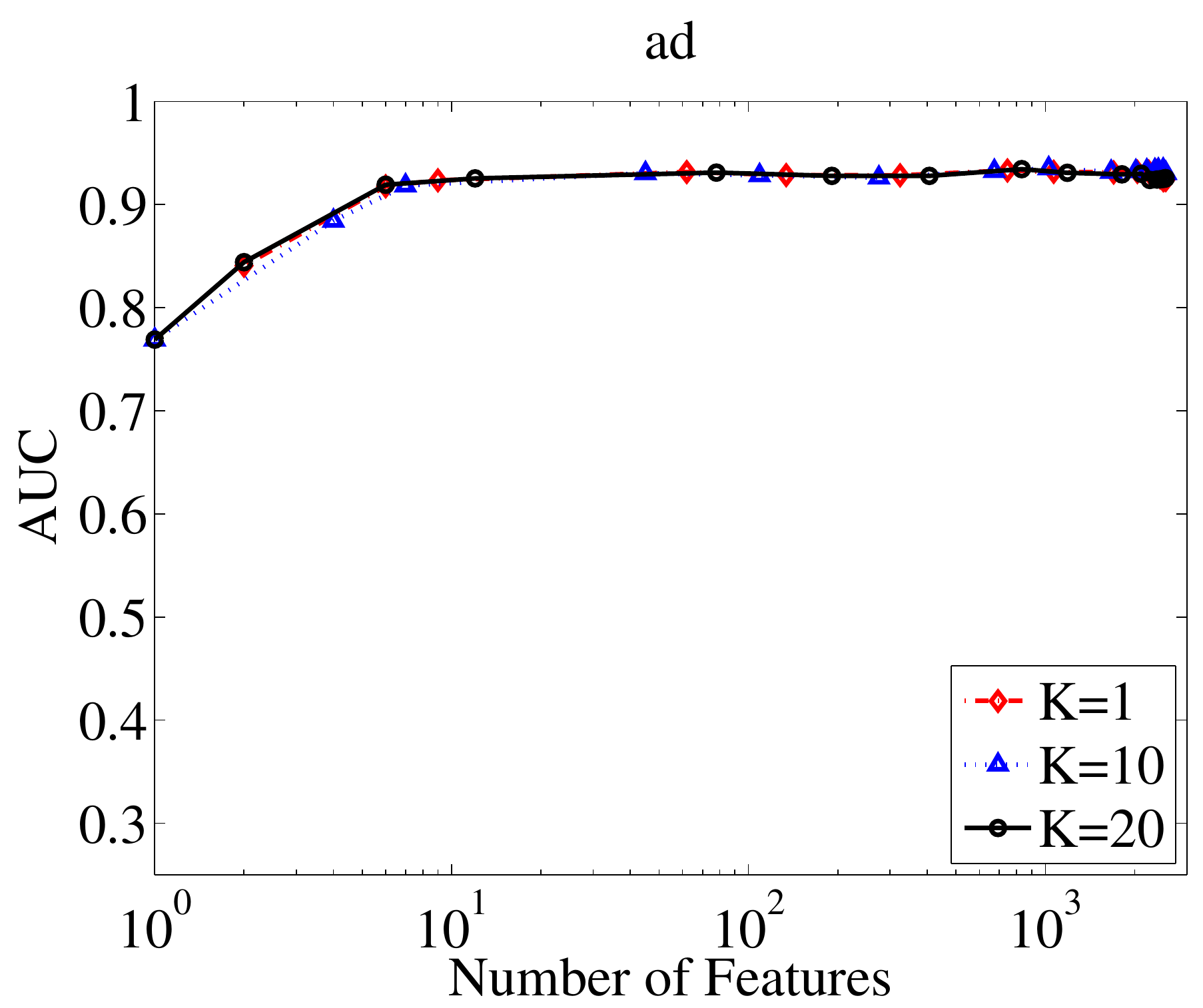}
&
\includegraphics[width=0.4\columnwidth]{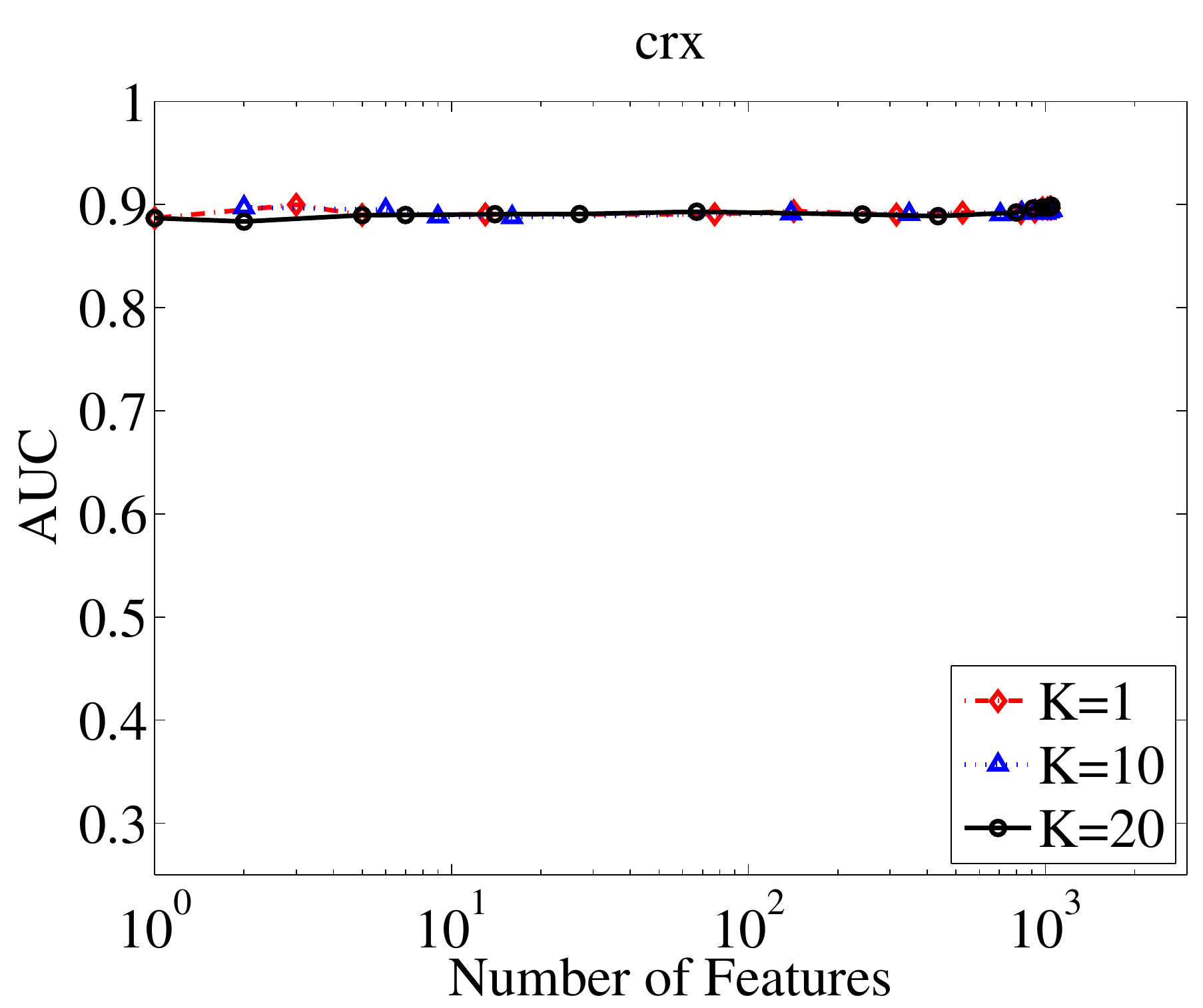}
\\
\includegraphics[width=0.4\columnwidth]{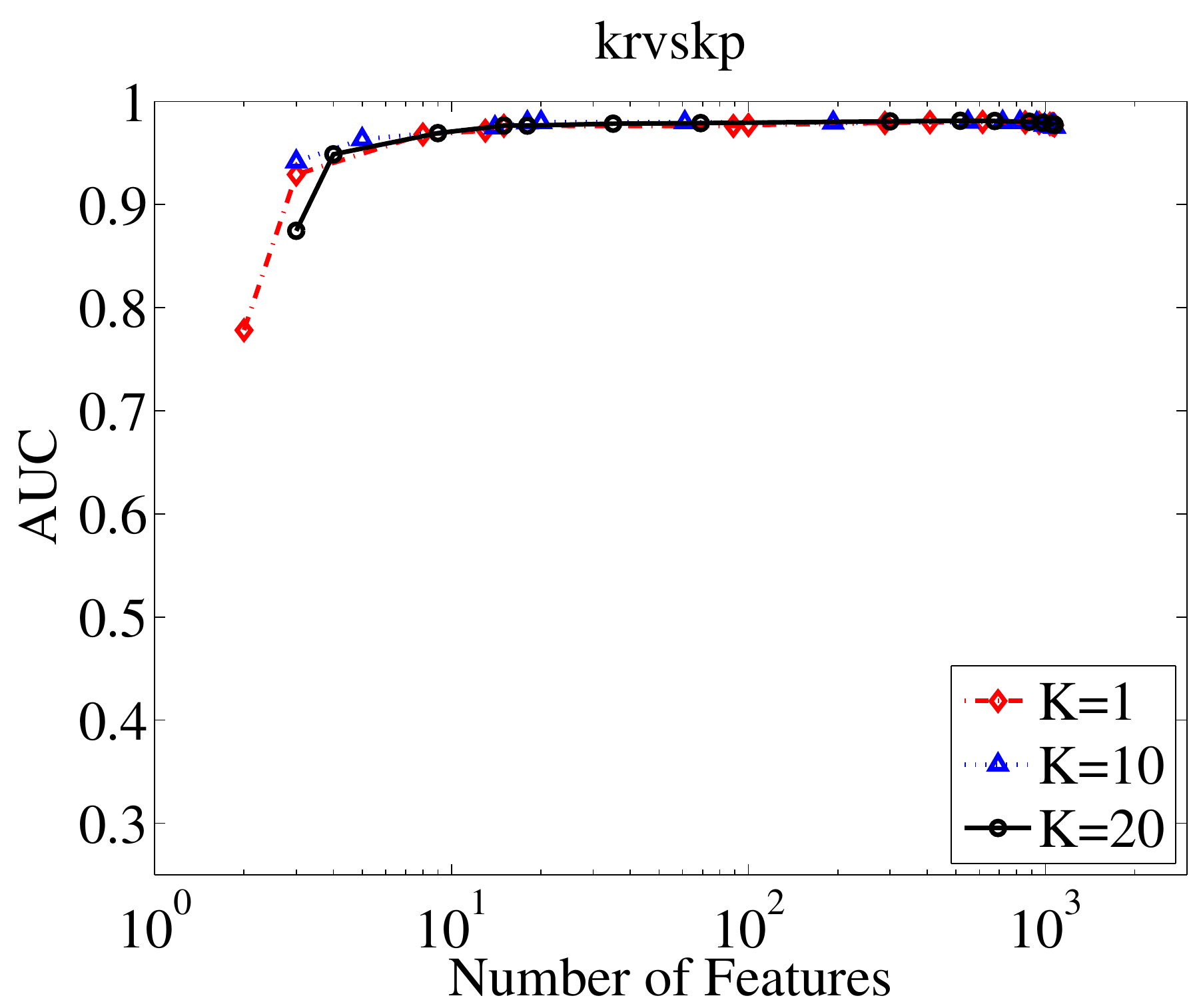}
&
\includegraphics[width=0.4\columnwidth]{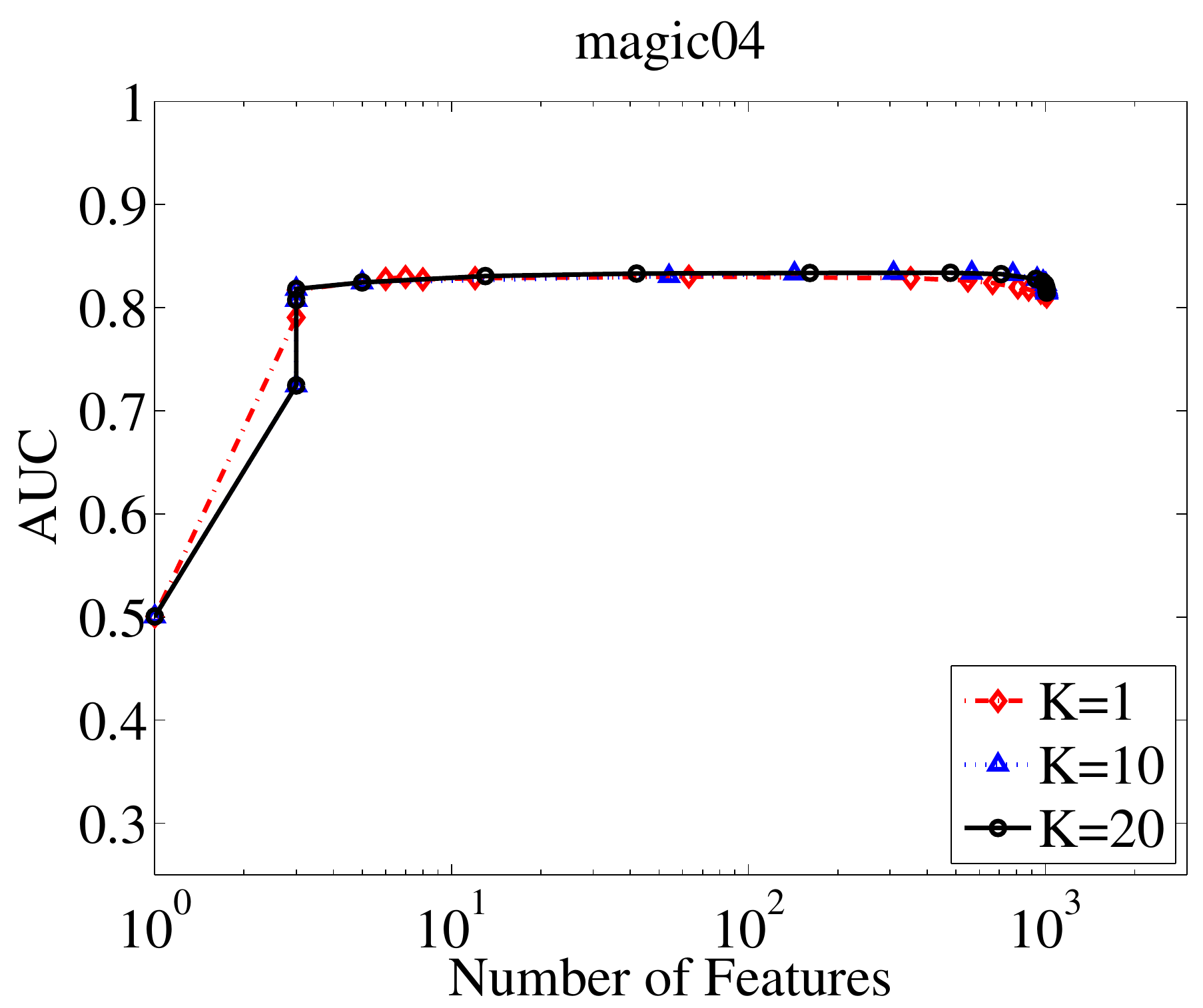}
\\
\includegraphics[width=0.4\columnwidth]{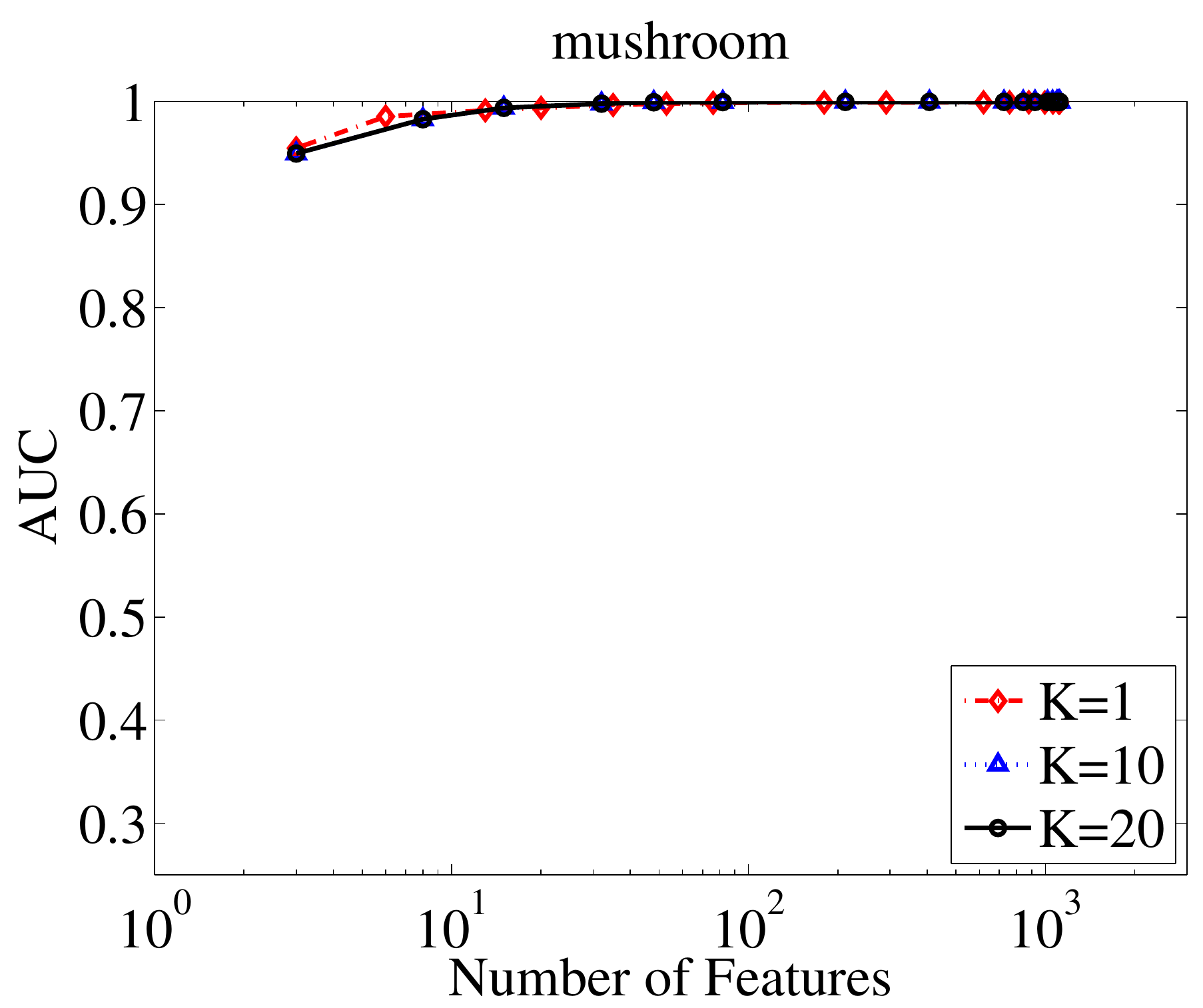}
&
\includegraphics[width=0.4\columnwidth]{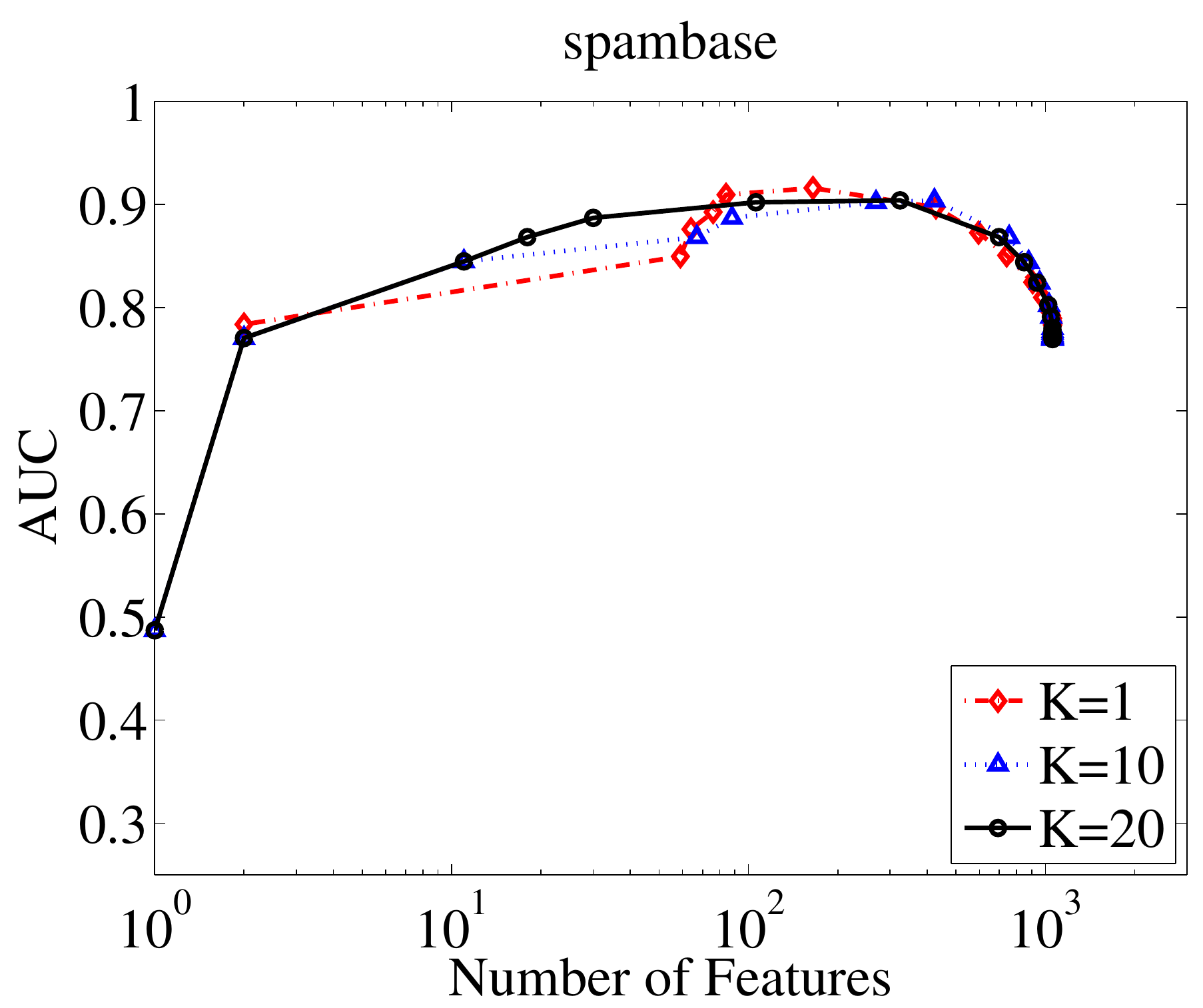}
\\
\includegraphics[width=0.4\columnwidth]{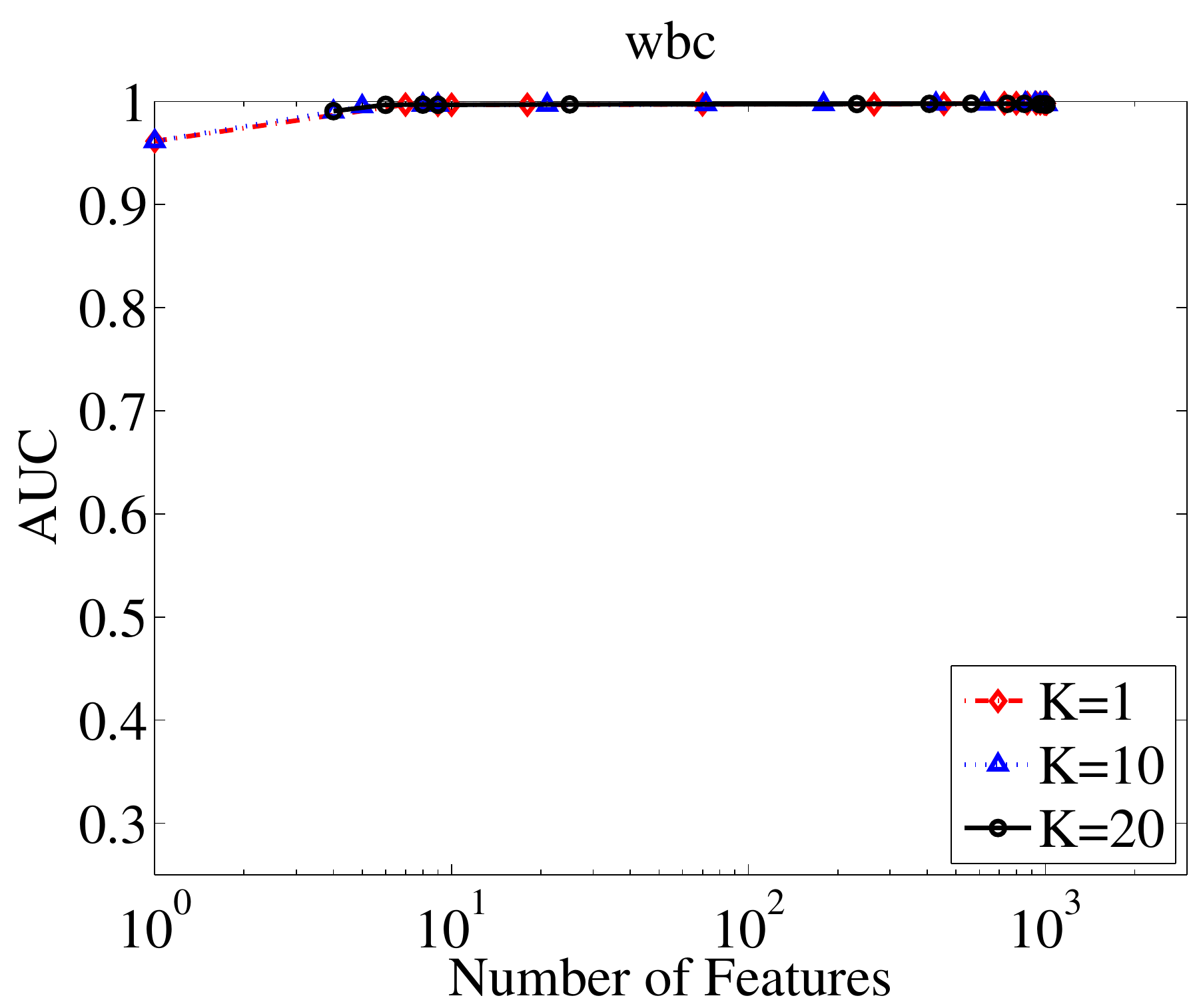}
&
\includegraphics[width=0.4\columnwidth]{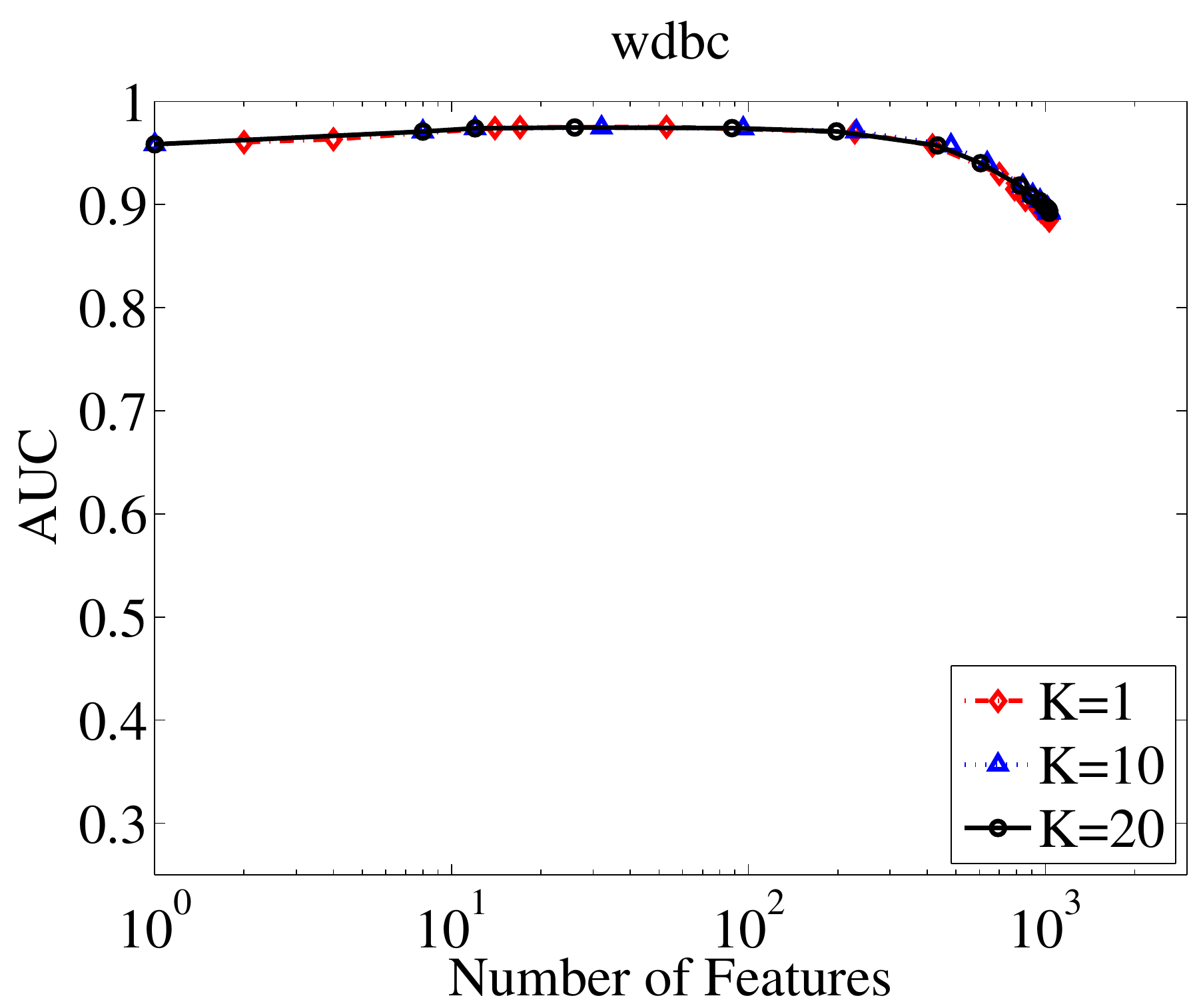}
\end{tabular}
\end{center}
\caption{Effect of $K$ on AUC in truncated gradient.}
\label{fig:k-tg2}
\end{figure}

\begin{figure}[t]
\begin{center}
\begin{tabular}{cc}
\includegraphics[width=0.4\columnwidth]{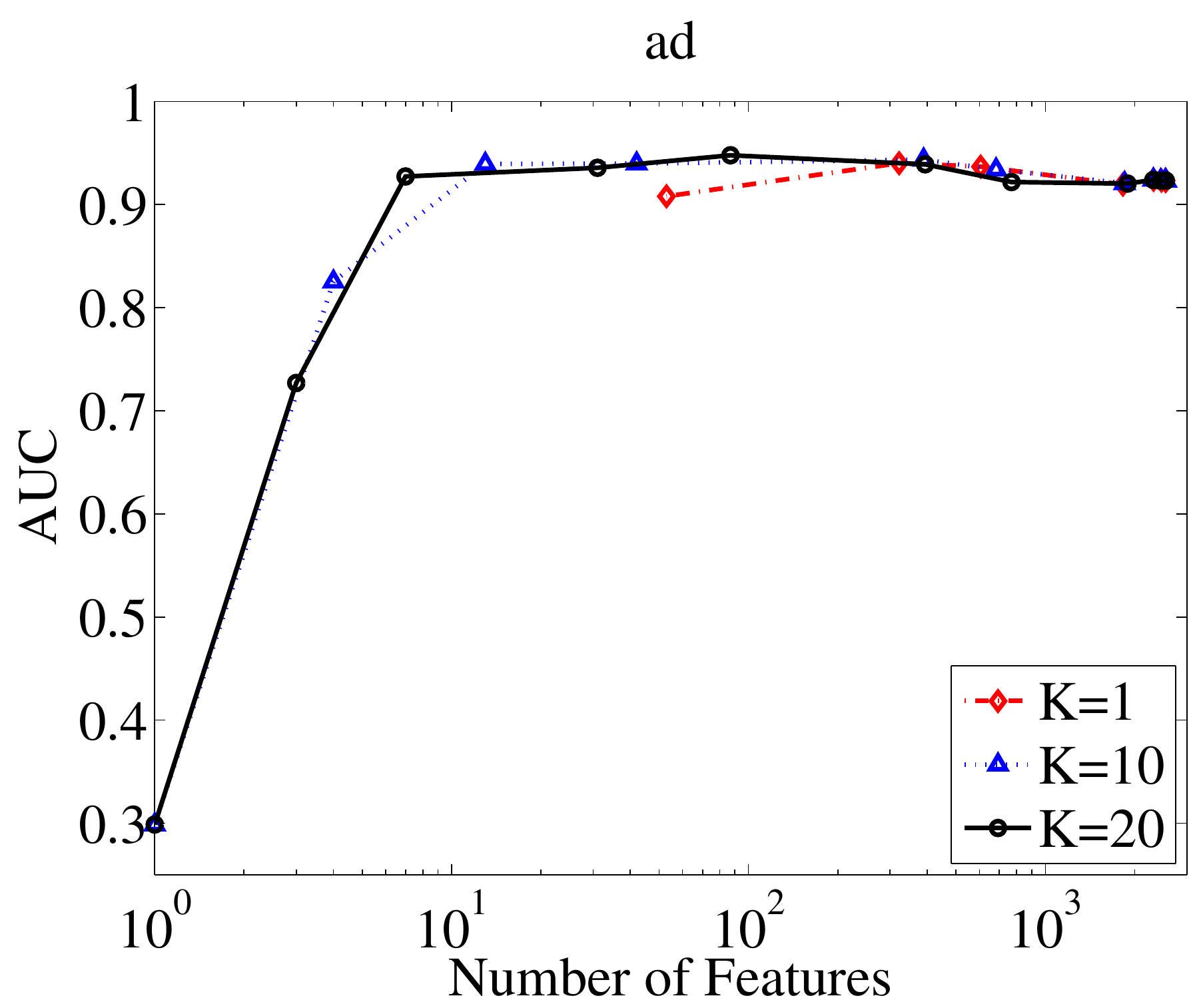}
&
\includegraphics[width=0.4\columnwidth]{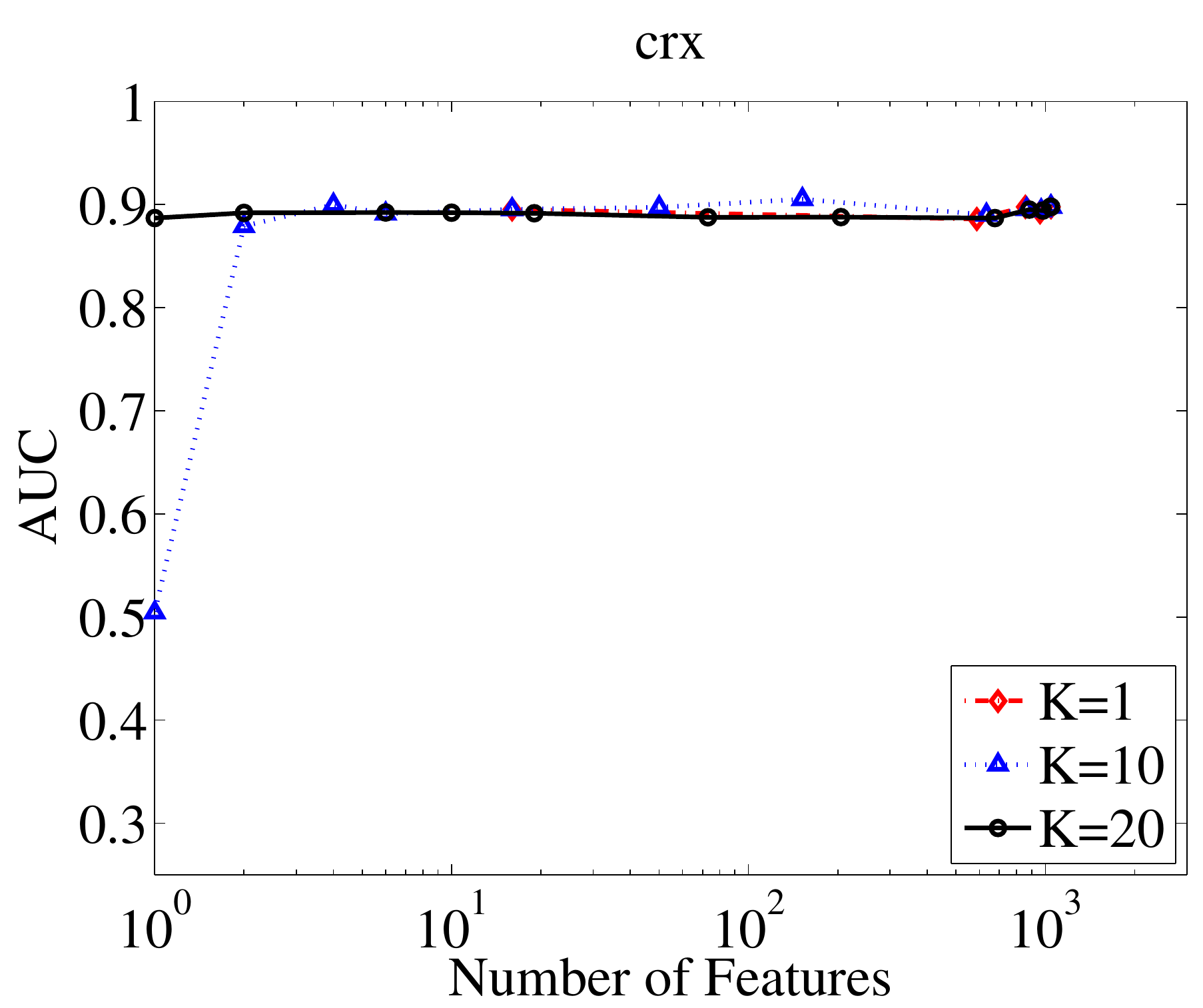}
\\
\includegraphics[width=0.4\columnwidth]{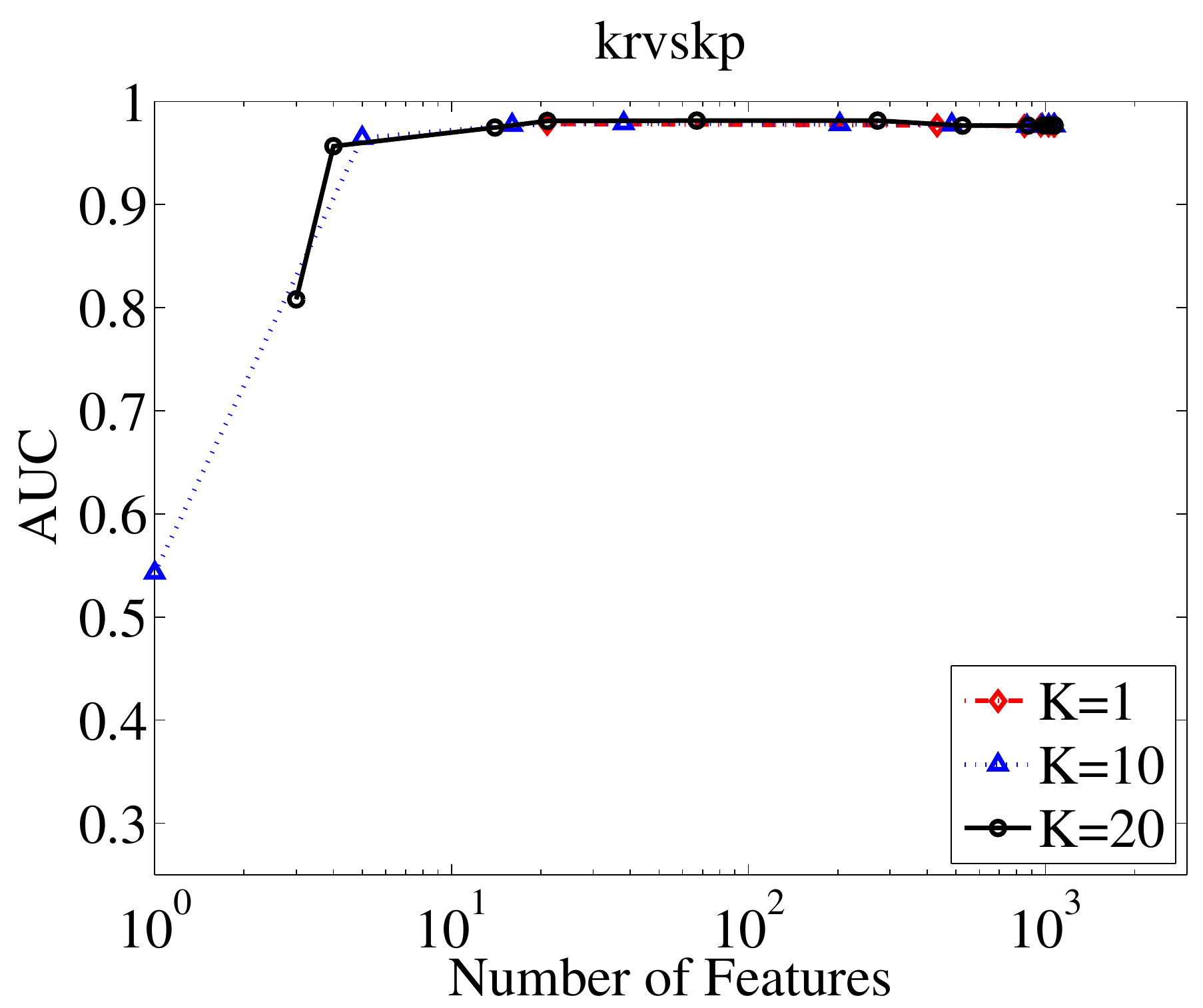}
&
\includegraphics[width=0.4\columnwidth]{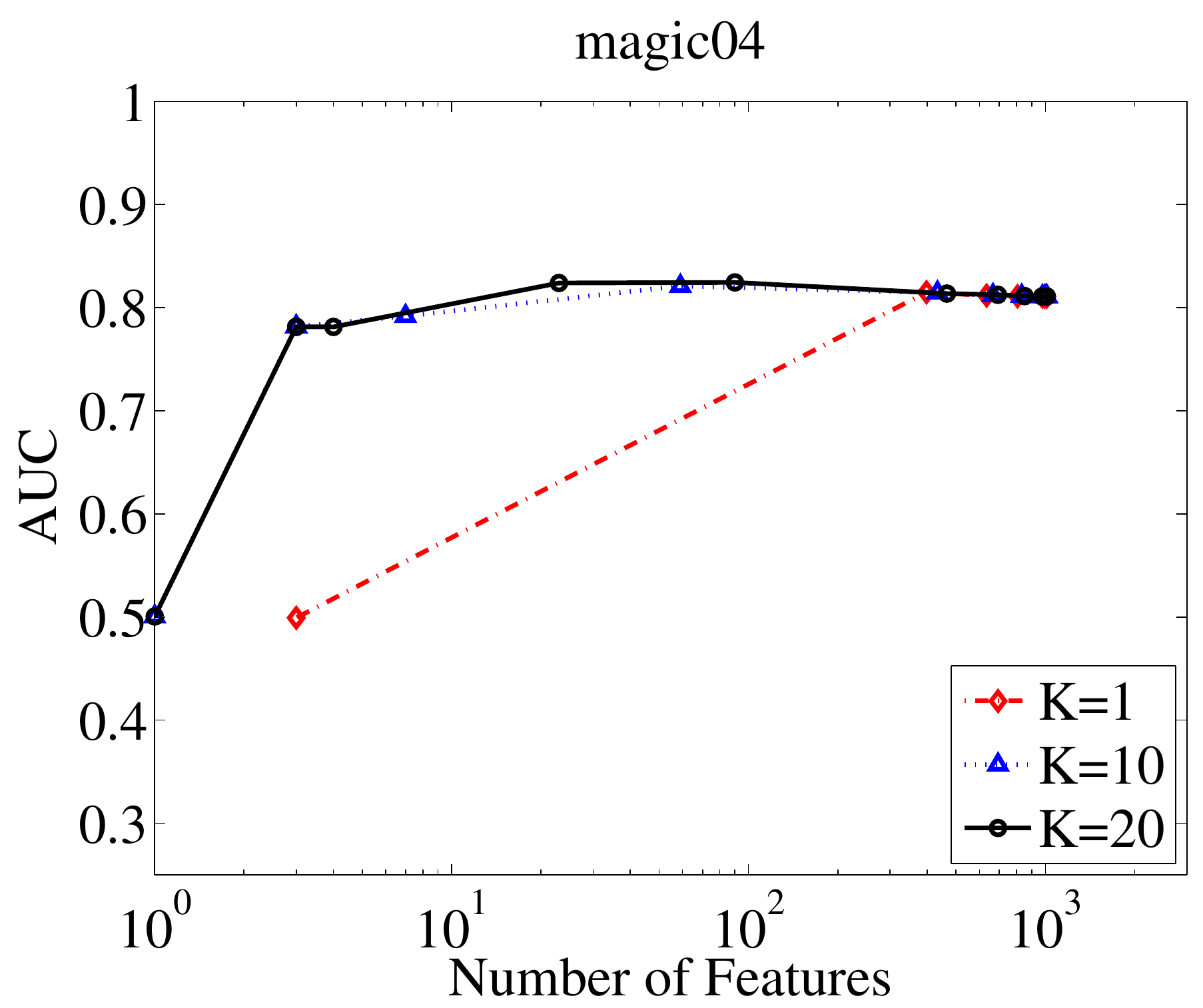}
\\
\includegraphics[width=0.4\columnwidth]{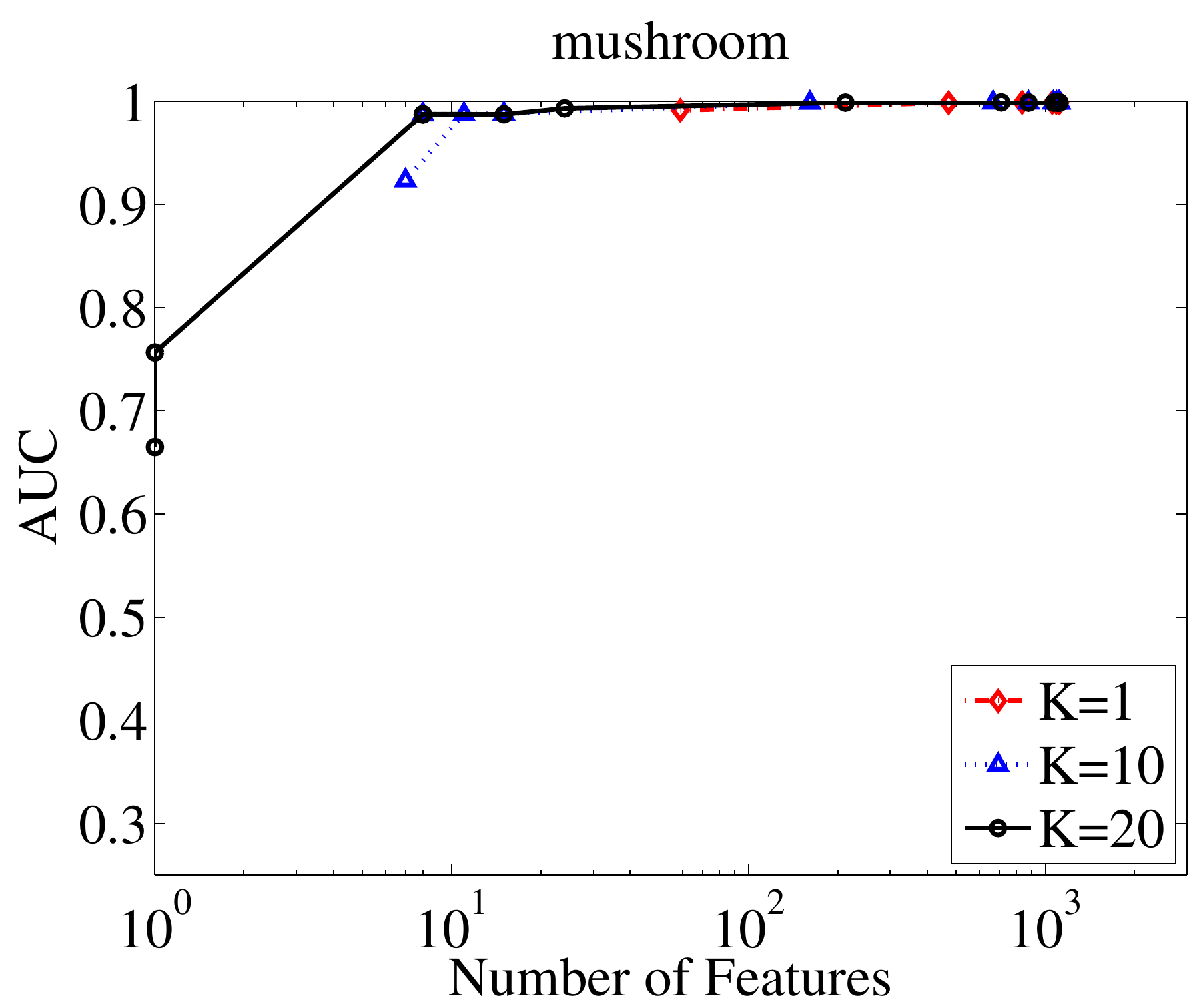}
&
\includegraphics[width=0.4\columnwidth]{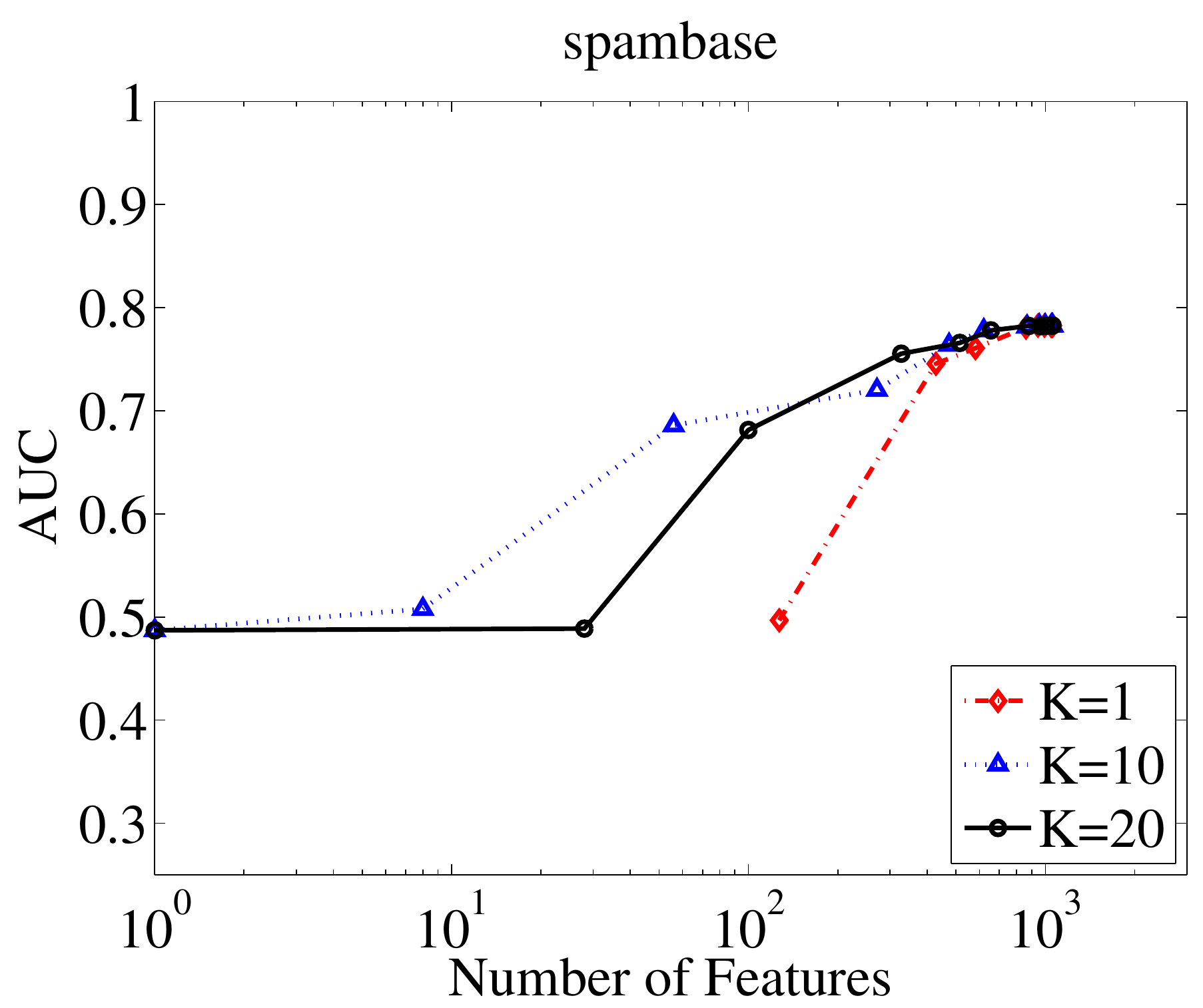}
\\
\includegraphics[width=0.4\columnwidth]{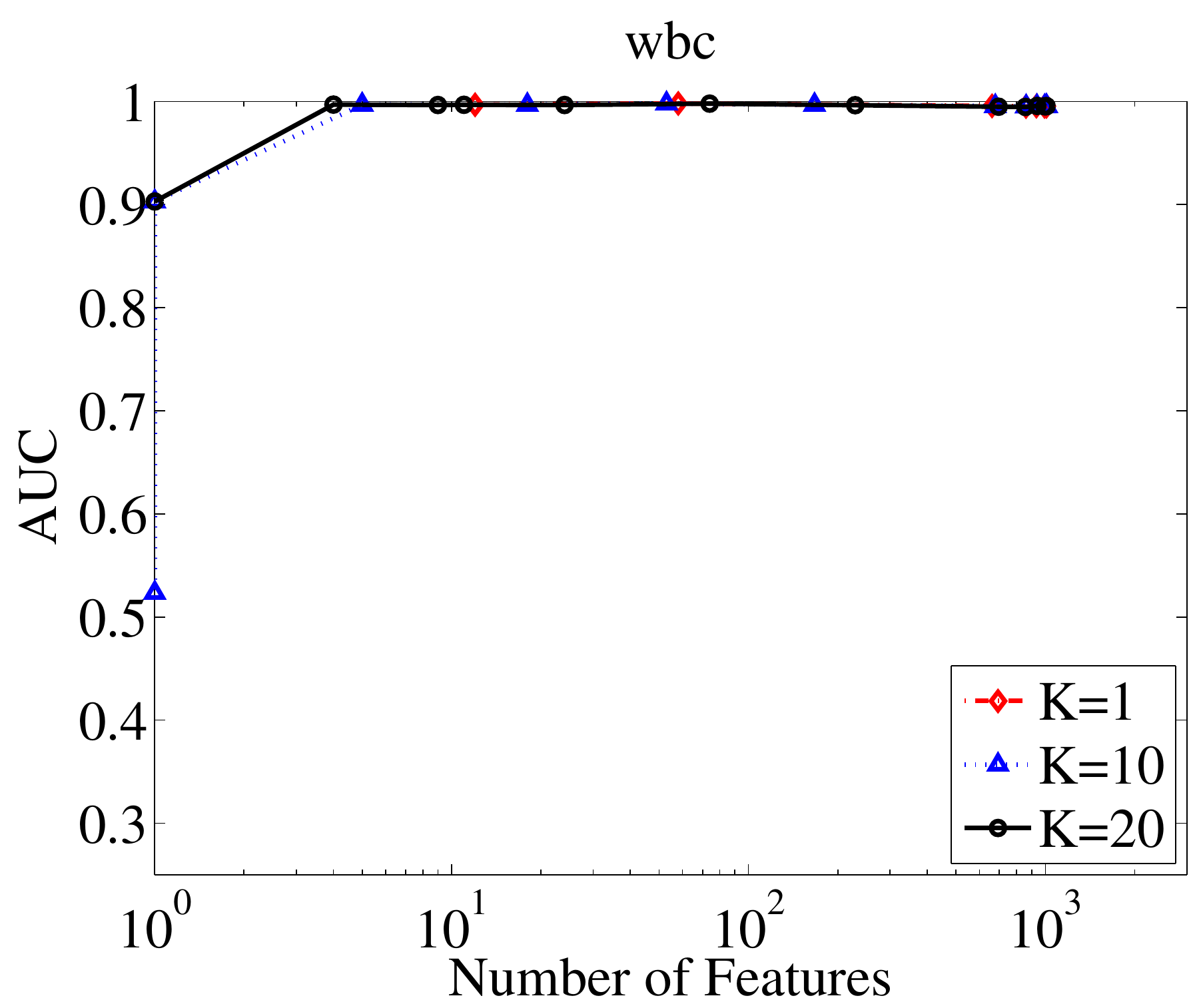}
&
\includegraphics[width=0.4\columnwidth]{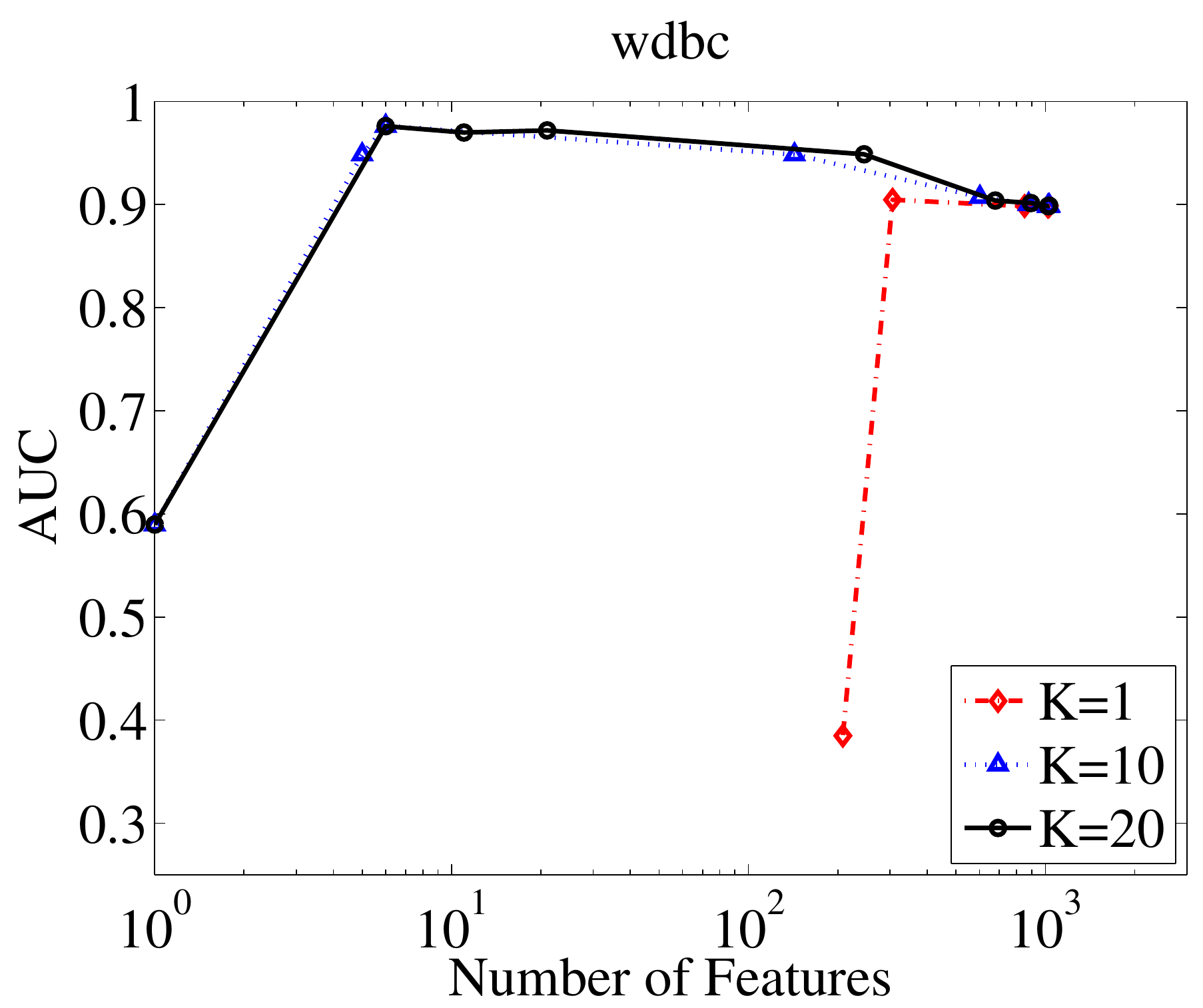}
\end{tabular}
\end{center}
\caption{Effect of $K$ on AUC in the rounding algorithm.}
\label{fig:k-r}
\end{figure}

Figures~\ref{fig:k-tg2} and \ref{fig:k-r} give the AUC vs.
number-of-feature plots, where each data point is generated by
running respective algorithm using a different value of $g$ (for
truncated gradient) and $\theta$ (for the rounding
algorithm). We used $\theta=\infty$ in truncated gradient.

For truncated gradient, the performances with $K=10$ or $20$ are
at least as good as those with $K=1$, and for the spambase dataset
further feature reduction is achieved at the same level of
performance, reducing the number of features from $76$ (when
$K=1$) to $25$ (when $K=10$ or $20$) with of an AUC of about
$0.89$.

Such an effect is even more remarkable in the rounding algorithm.
For instance, in the ad dataset the algorithm using $K=1$ achieves
an AUC of $0.94$ with $322$ features, while $13$ and $7$ features
are needed using $K=10$ and $K=20$, respectively.

\subsection{The Effects of $\theta$ in Truncated Gradient}

In this subsection, we empirically study the effect of 
$\theta$ in truncated gradient.  The rounding algorithm
is also included for comparison due to its similarity to truncated
gradient when $\theta=g$.  As before, we used cross validation to
choose parameters for each $\theta$ value tried, and focused on
the AUC metric in the eight UCI classification tasks, except the
degenerate one of wpbc.  We fixed $K=10$ in both algorithm.

\begin{figure}[t]
\begin{center}
\begin{tabular}{cc}
\includegraphics[width=0.4\columnwidth]{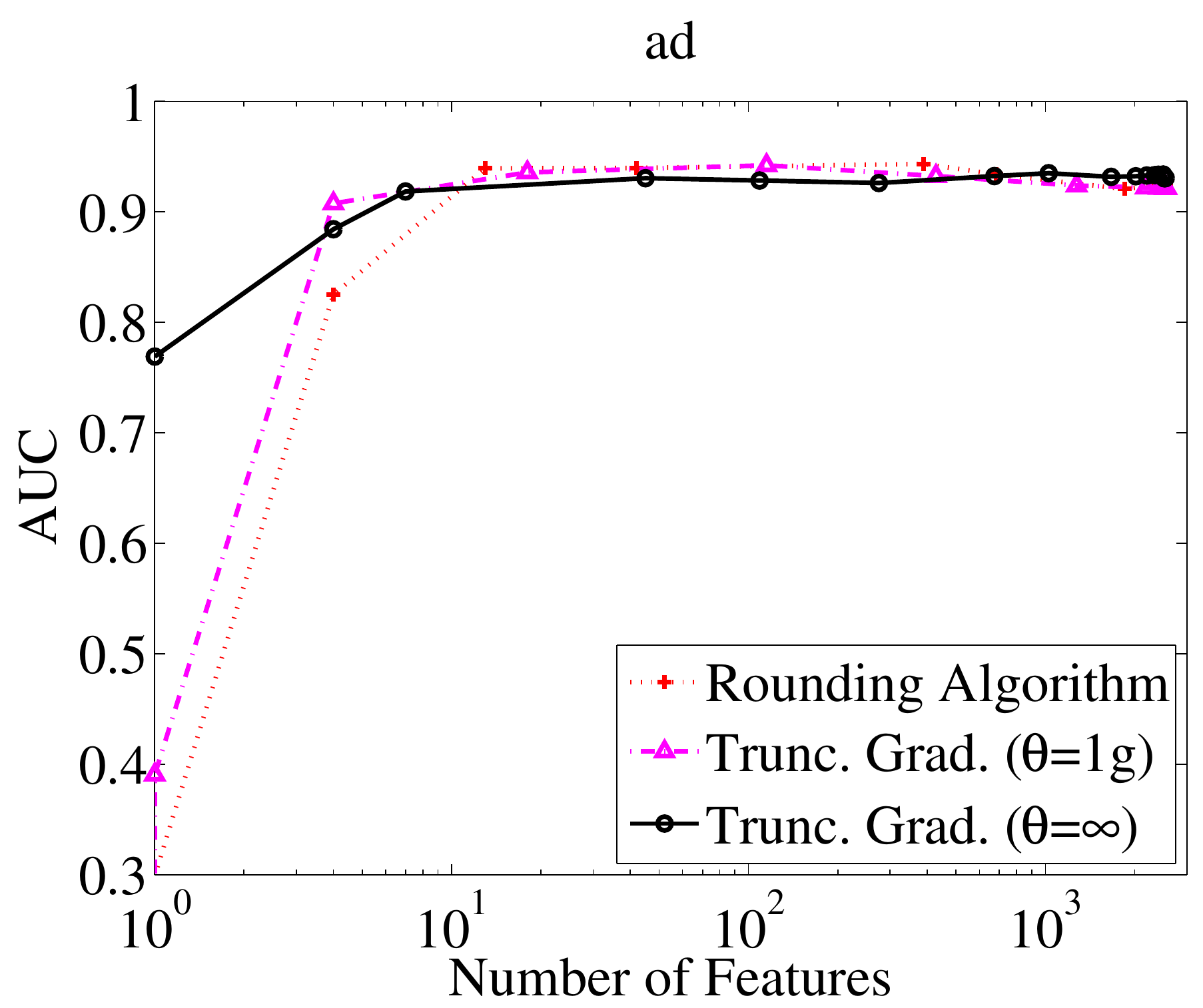}
&
\includegraphics[width=0.4\columnwidth]{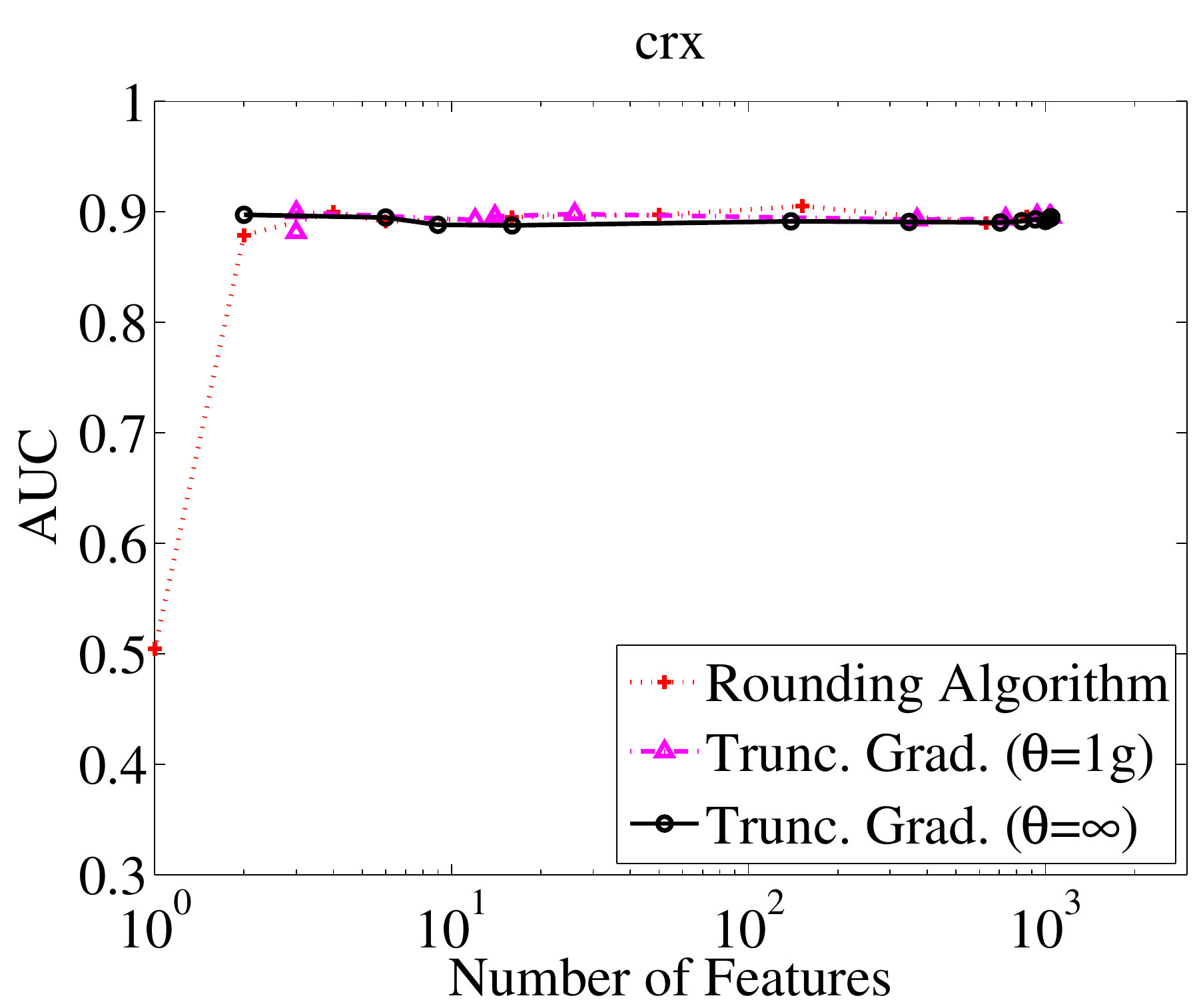}
\\
\includegraphics[width=0.4\columnwidth]{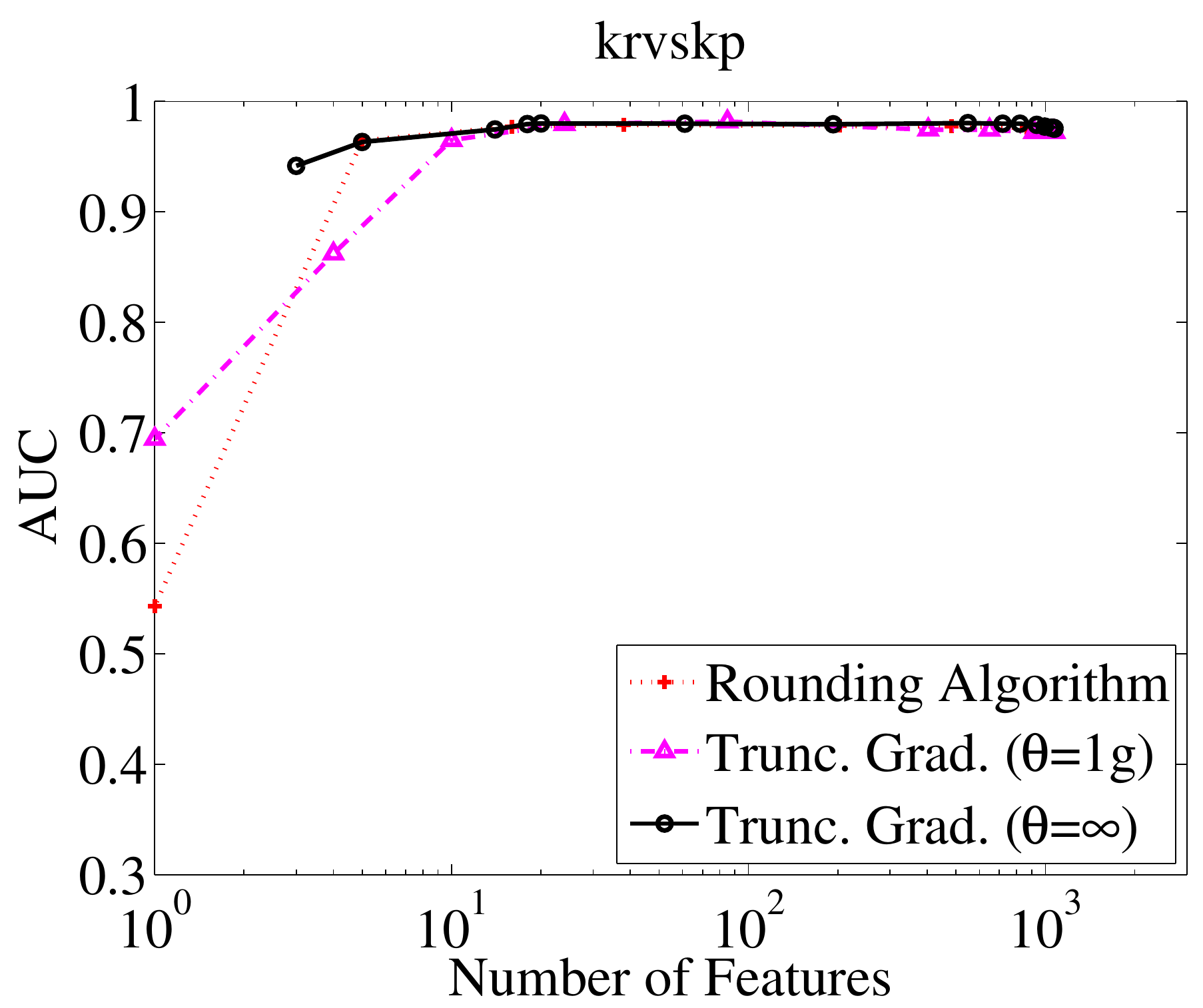}
&
\includegraphics[width=0.4\columnwidth]{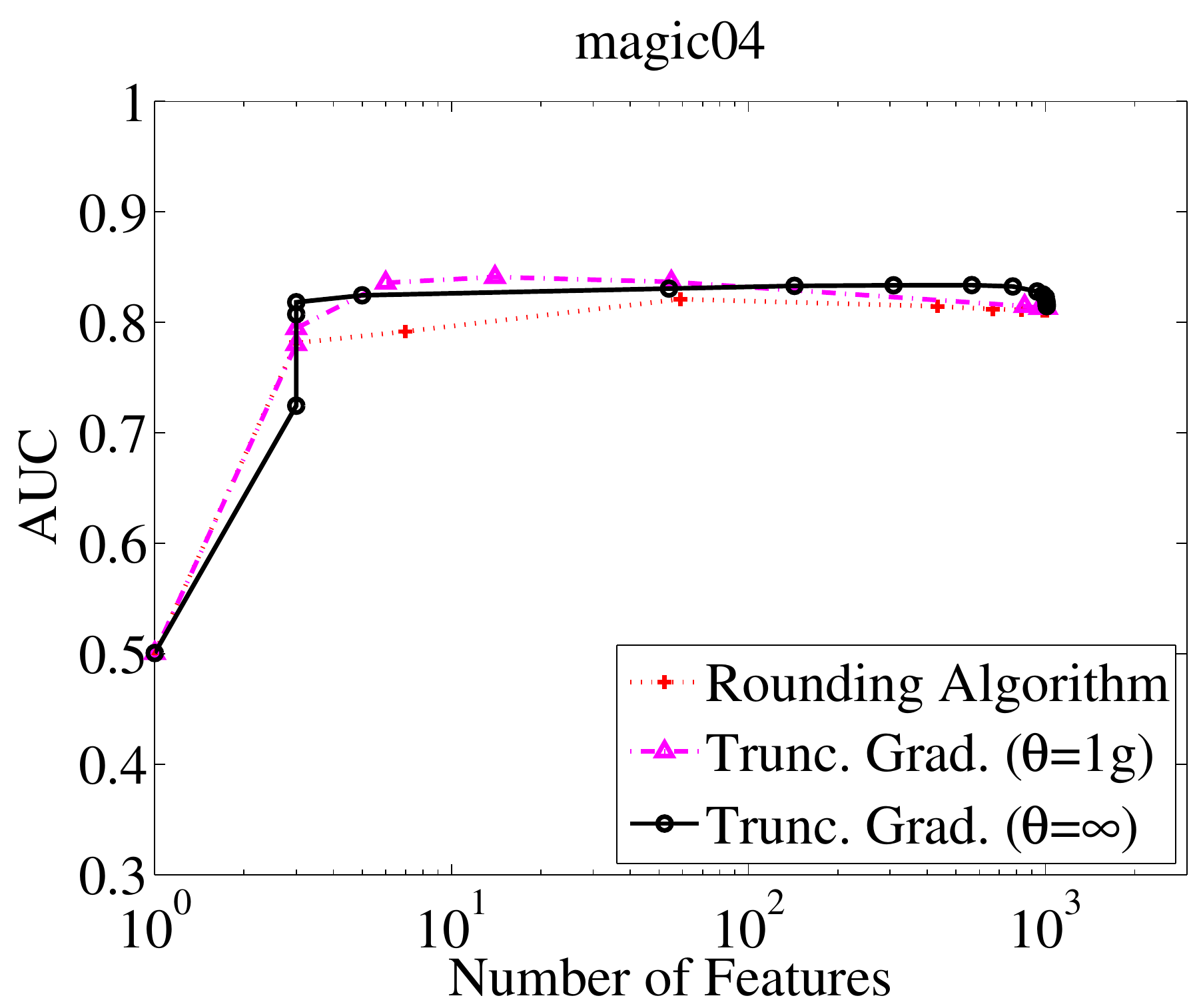}
\\
\includegraphics[width=0.4\columnwidth]{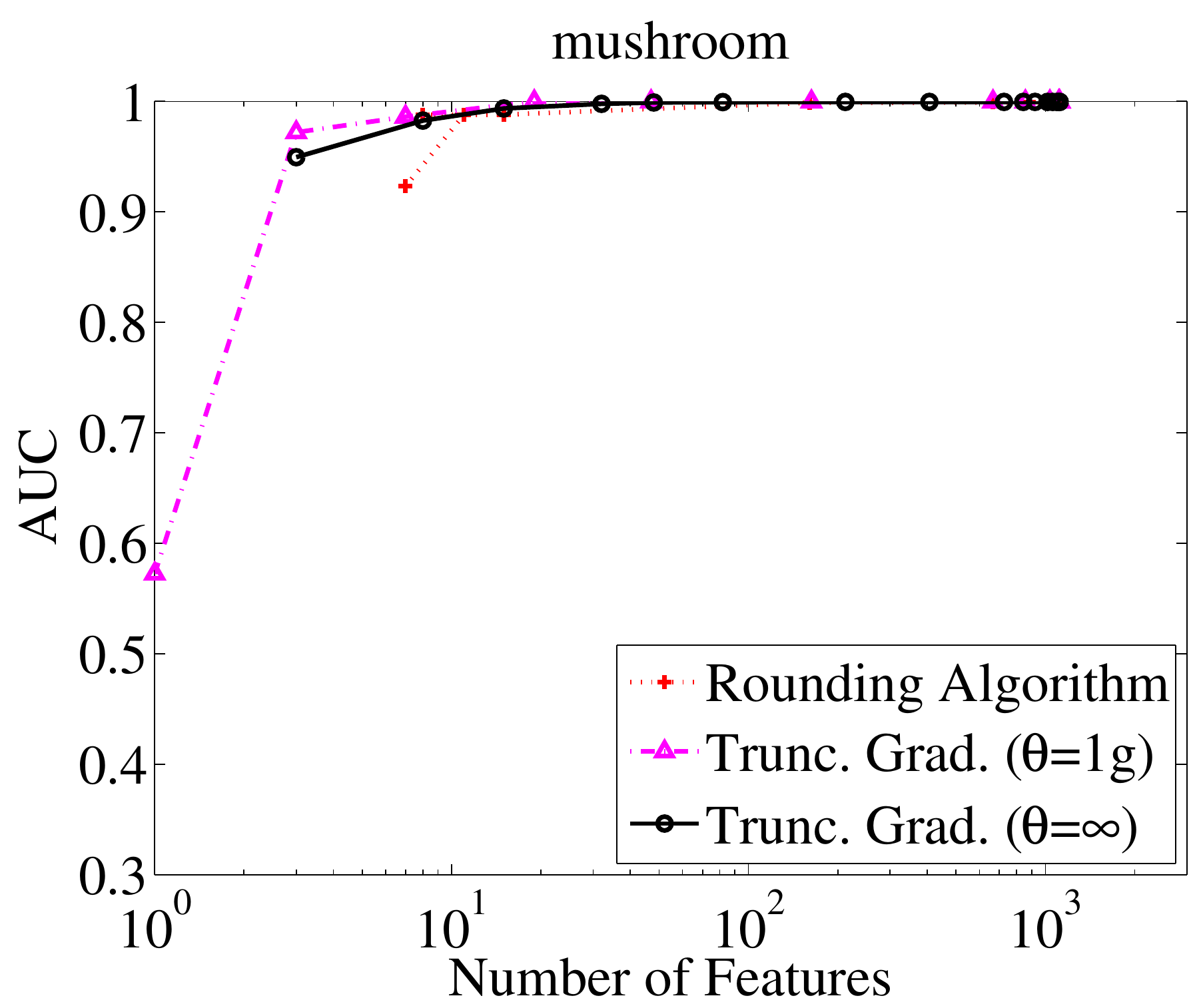}
&
\includegraphics[width=0.4\columnwidth]{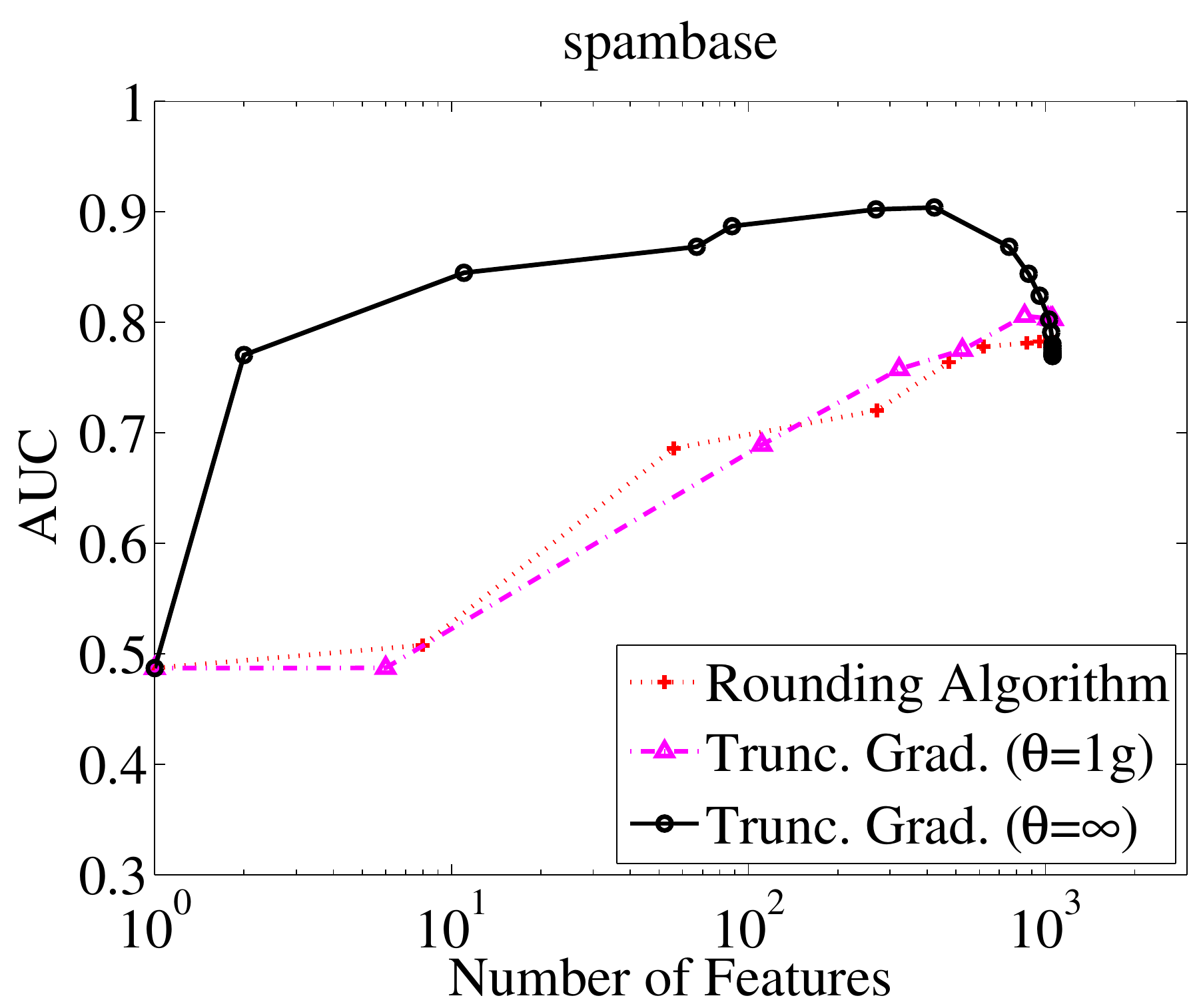}
\\
\includegraphics[width=0.4\columnwidth]{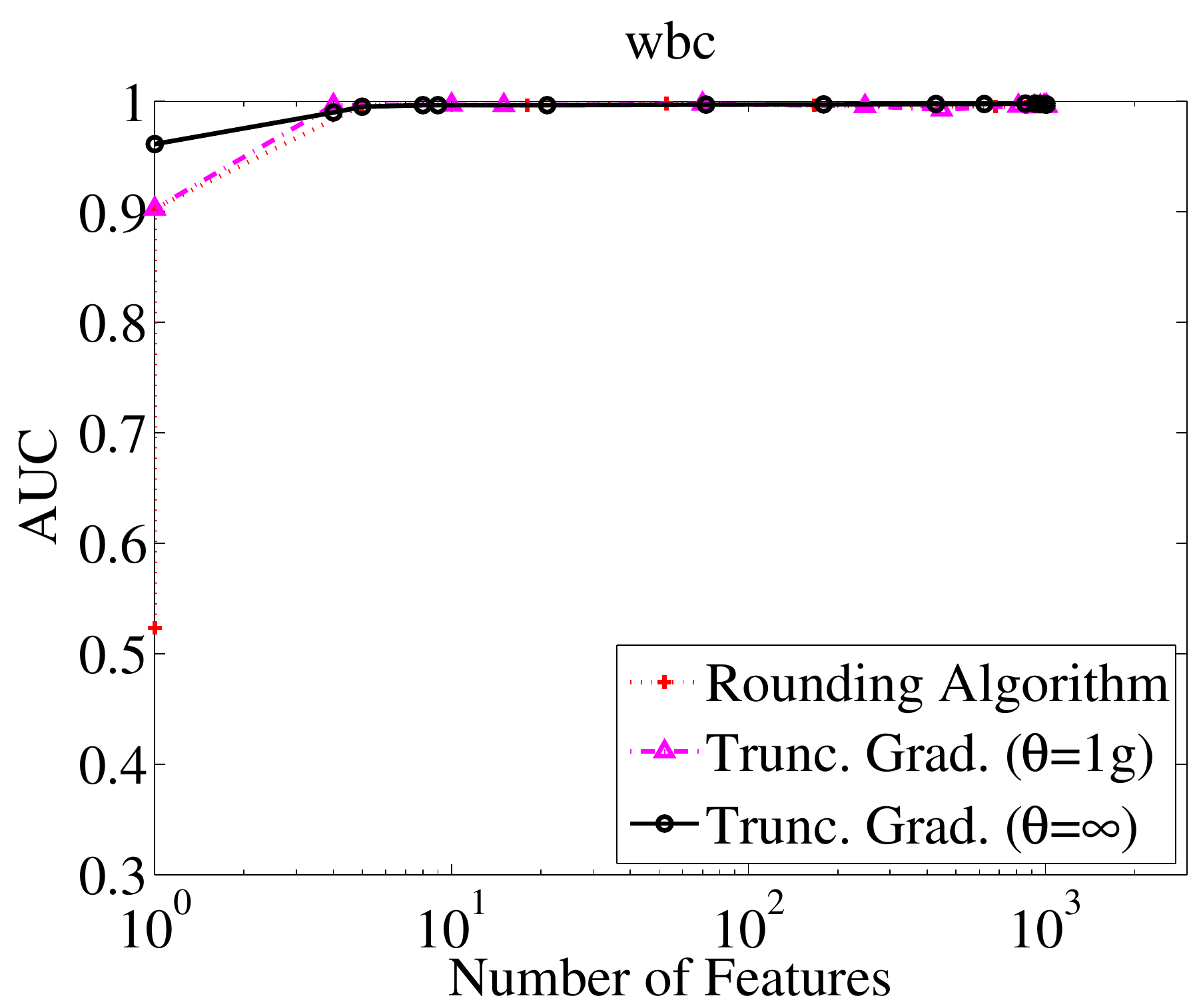}
&
\includegraphics[width=0.4\columnwidth]{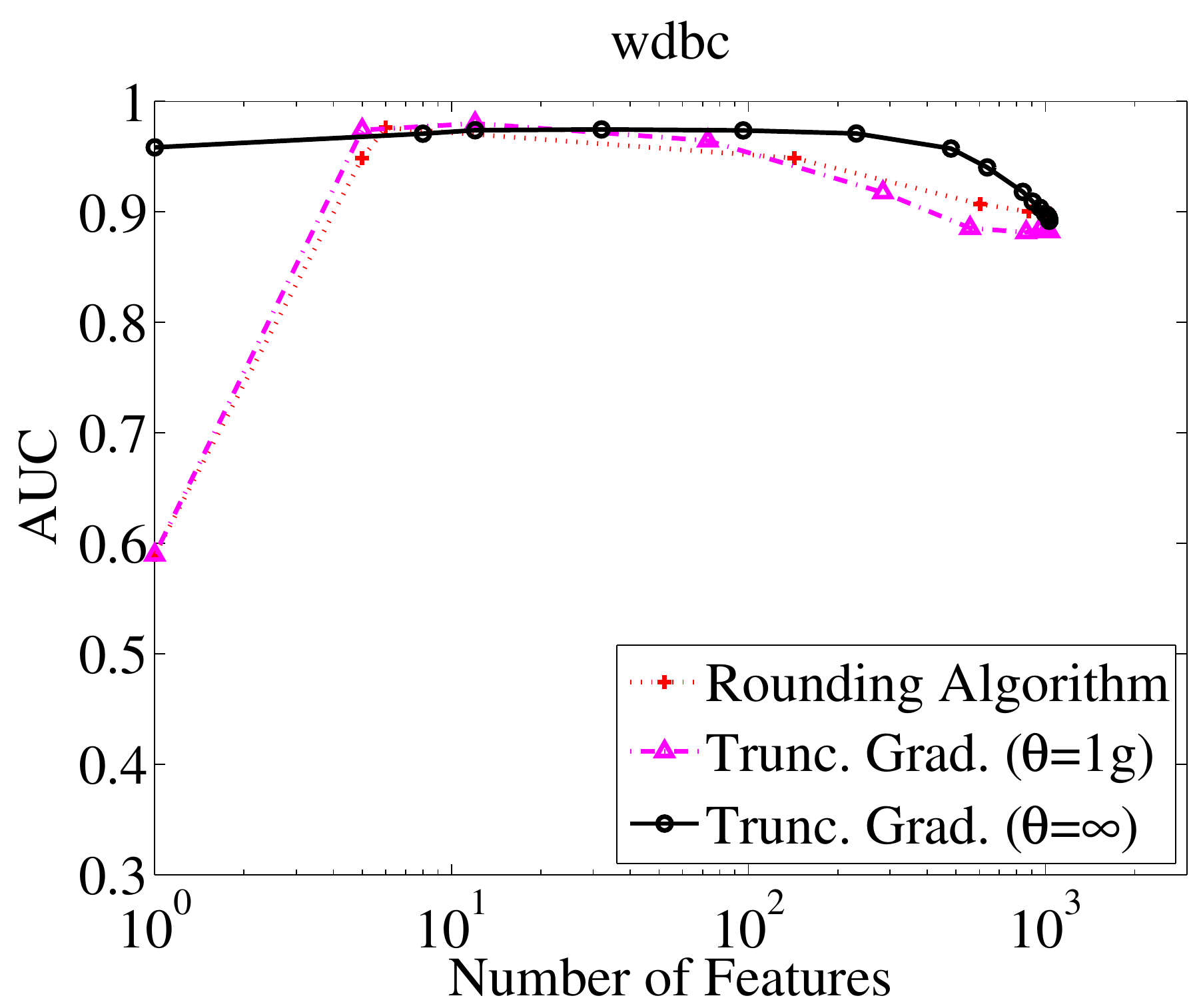}
\end{tabular}
\end{center}
\caption{Effect of $\theta$ on AUC in truncated gradient.}
\label{fig:theta-effect}
\end{figure}

Figure~\ref{fig:theta-effect} gives the AUC vs. number-of-feature
plots, where each data point is generated by running respective
algorithms using a different value of $g$ (for truncated gradient)
and $\theta$ (for the rounding algorithm).  A few observations are
in place.  First, the results verify the observation that the
behavior of truncated gradient with $\theta=g$ is similar to the
rounding algorithm. 
Second, these results suggest that, in
practice, it may be desired to use $\theta=\infty$ in truncated
gradient because it avoids the local minimum problem.

\subsection{Comparison to Other Algorithms}

The next set of experiments compares truncated gradient descent to
other algorithms regarding their abilities to tradeoff feature
sparsification and performance.  Again, we focus on the AUC metric
in UCI classification tasks except wpdc.  The algorithms for
comparison include:
\begin{itemize}
\item{The truncated gradient algorithm: We fixed $K=10$ and
$\theta=\infty$, used crossed-validated parameters, and altered
the gravity parameter $g$.}
\item{The rounding algorithm described in
section~\ref{sec:rounding}: We fixed $K=10$, used cross-validated
parameters, and altered the rounding threshold $\theta$.}
\item{The subgradient algorithm described in
section~\ref{sec:subgradient}: We fixed $K=10$, used
cross-validated parameters, and altered the regularization
parameter $g$.}
\item{The Lasso~\cite{Lasso} for batch $L_1$ regularization: We
used a publicly available implementation~\cite{LassoSoftware}.}
\end{itemize}
Note that we do not attempt to compare these algorithms on rcv1
and \bigads simply because their sizes are too large for the Lasso
and subgradient descent (\textit{c.f.},
section~\ref{sec:implementation}).

\begin{figure}[t]
\begin{center}
\begin{tabular}{cc}
\includegraphics[width=0.4\columnwidth]{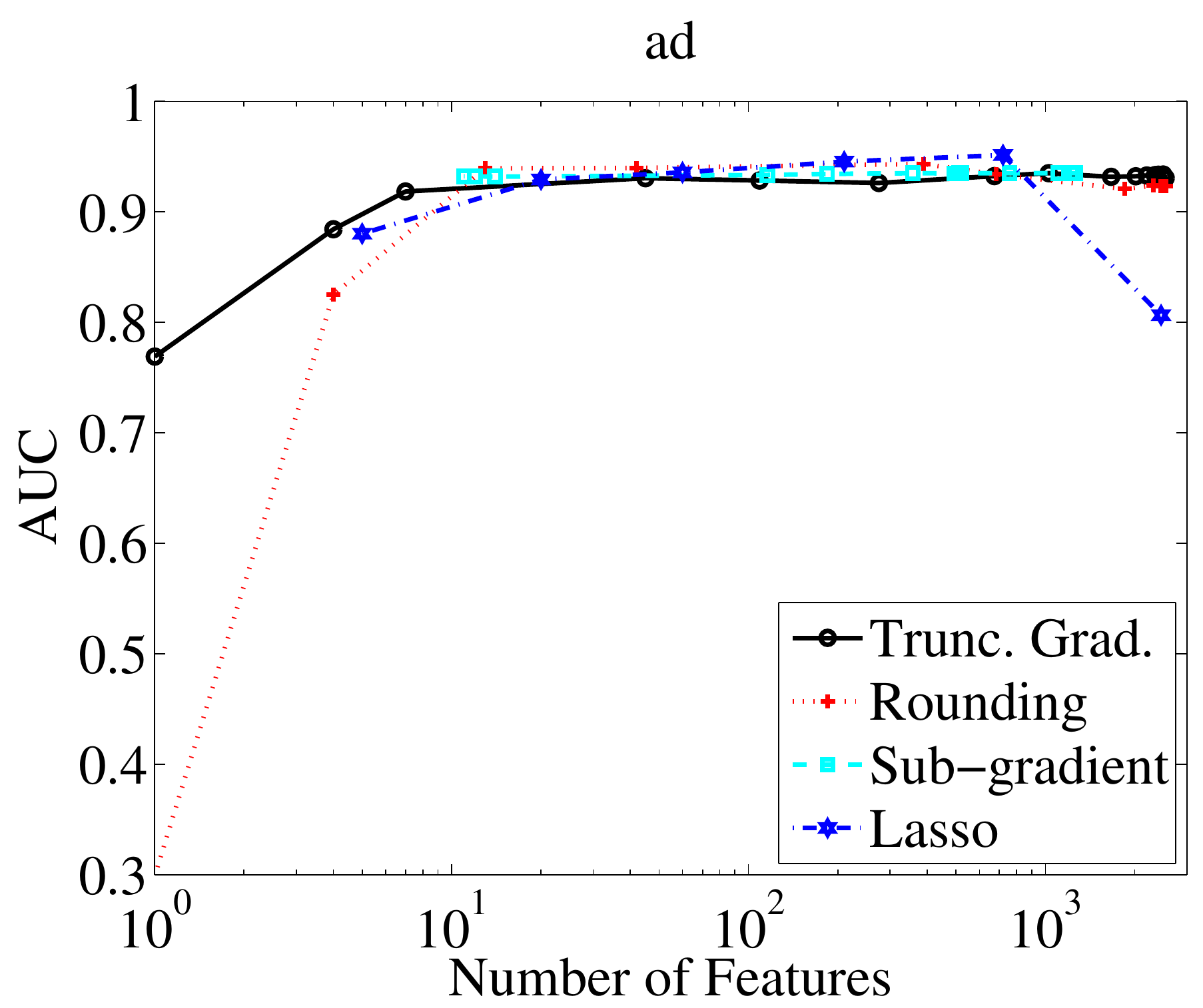}
&
\includegraphics[width=0.4\columnwidth]{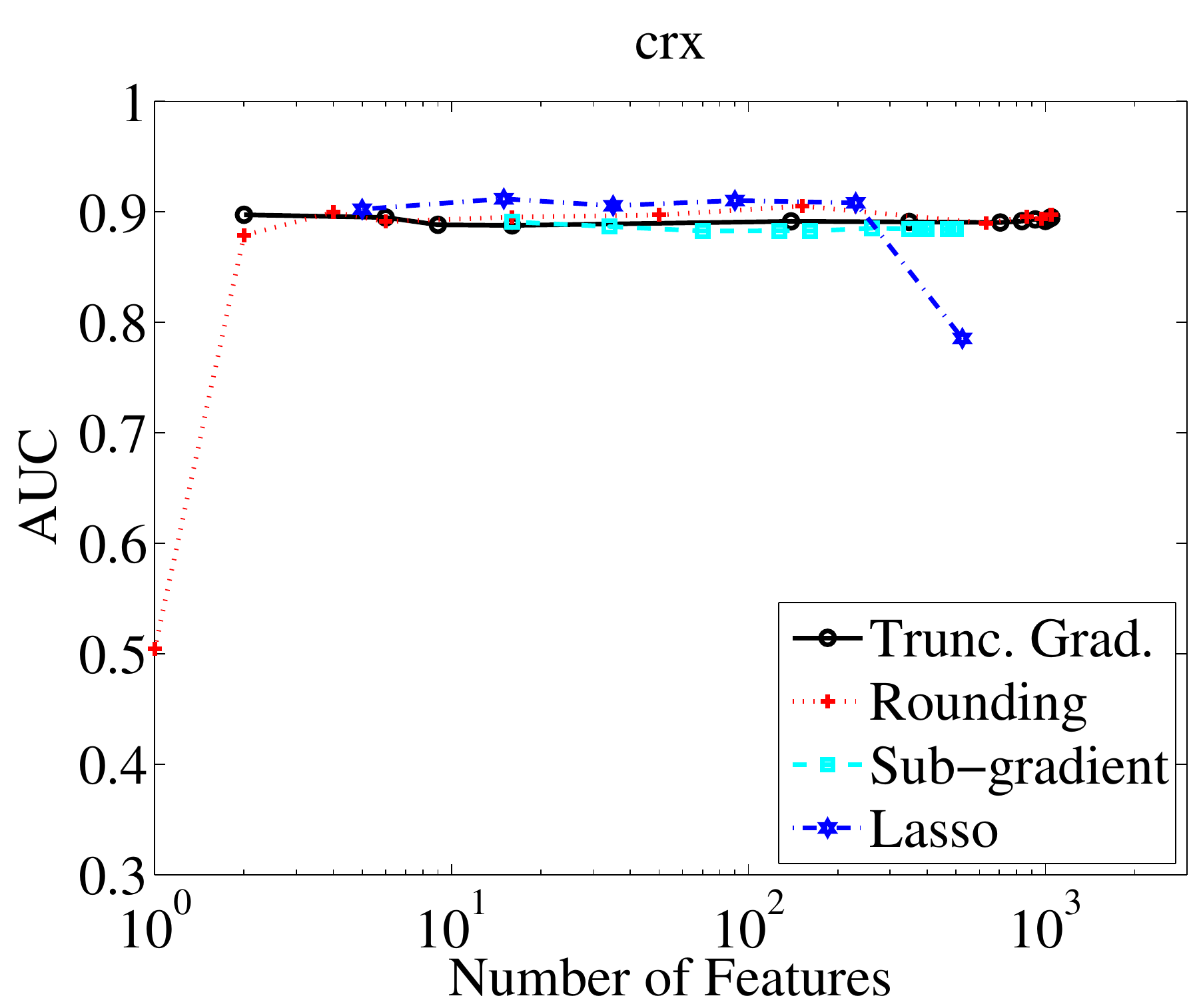}
\\
\includegraphics[width=0.4\columnwidth]{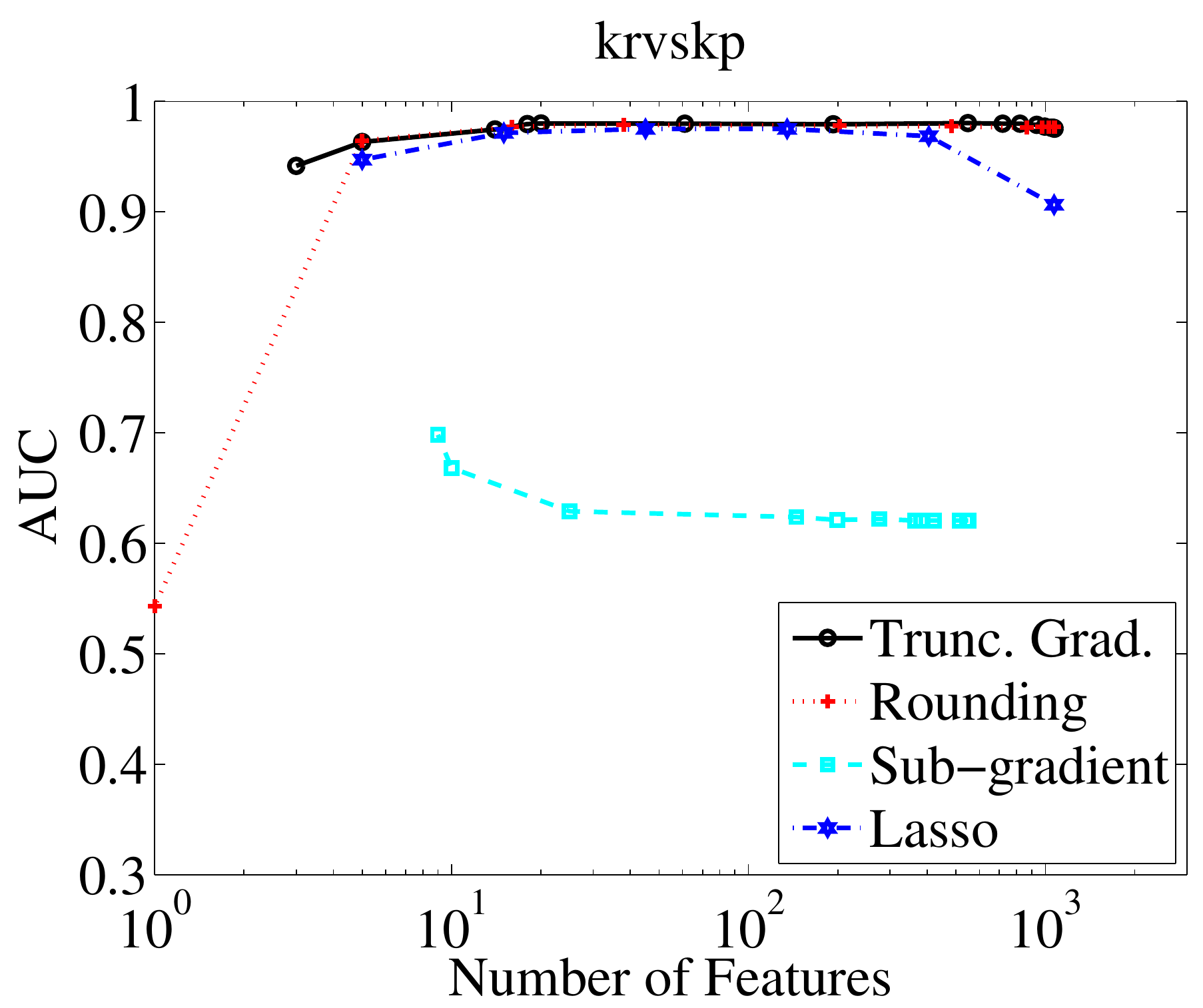}
&
\includegraphics[width=0.4\columnwidth]{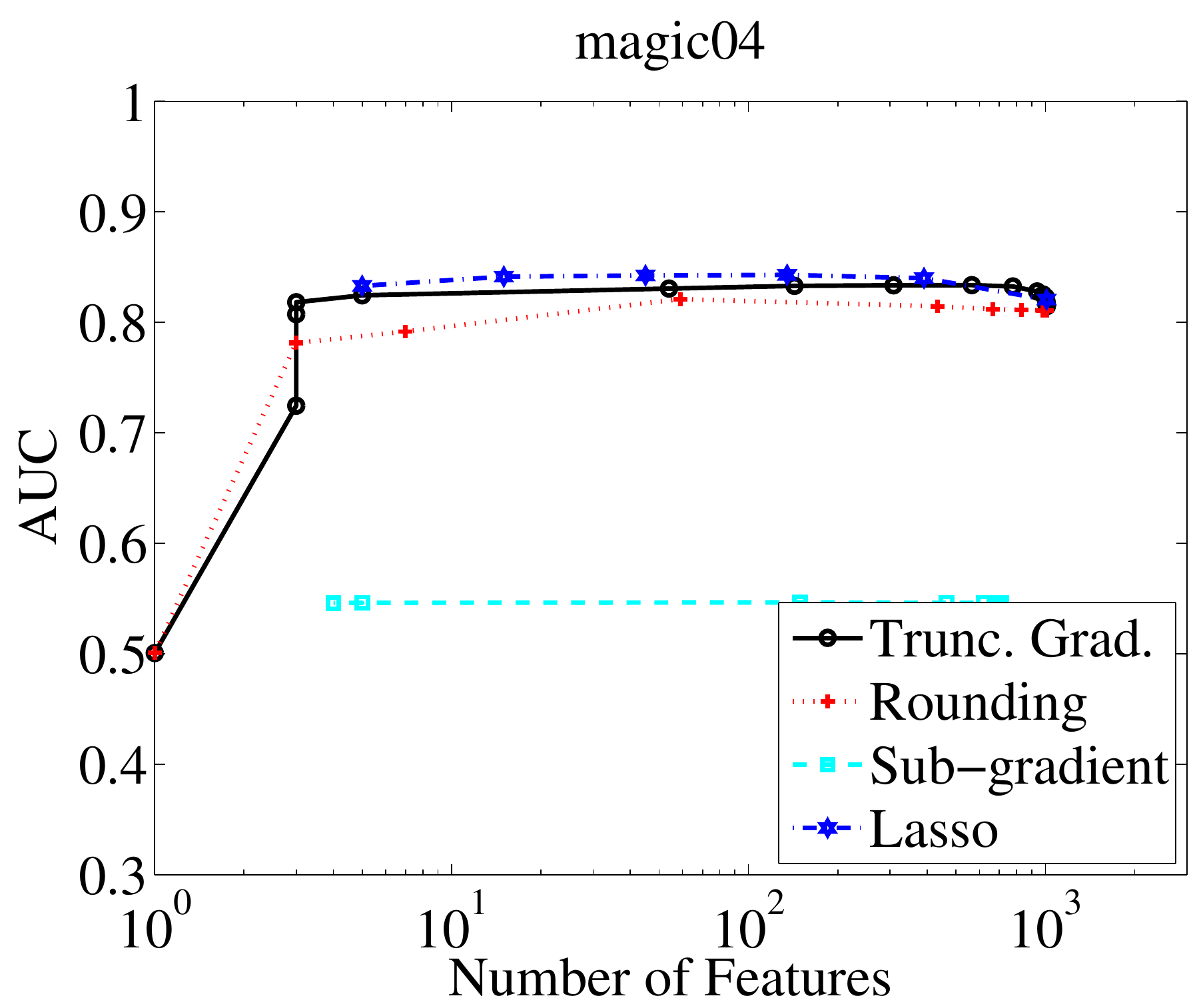}
\\
\includegraphics[width=0.4\columnwidth]{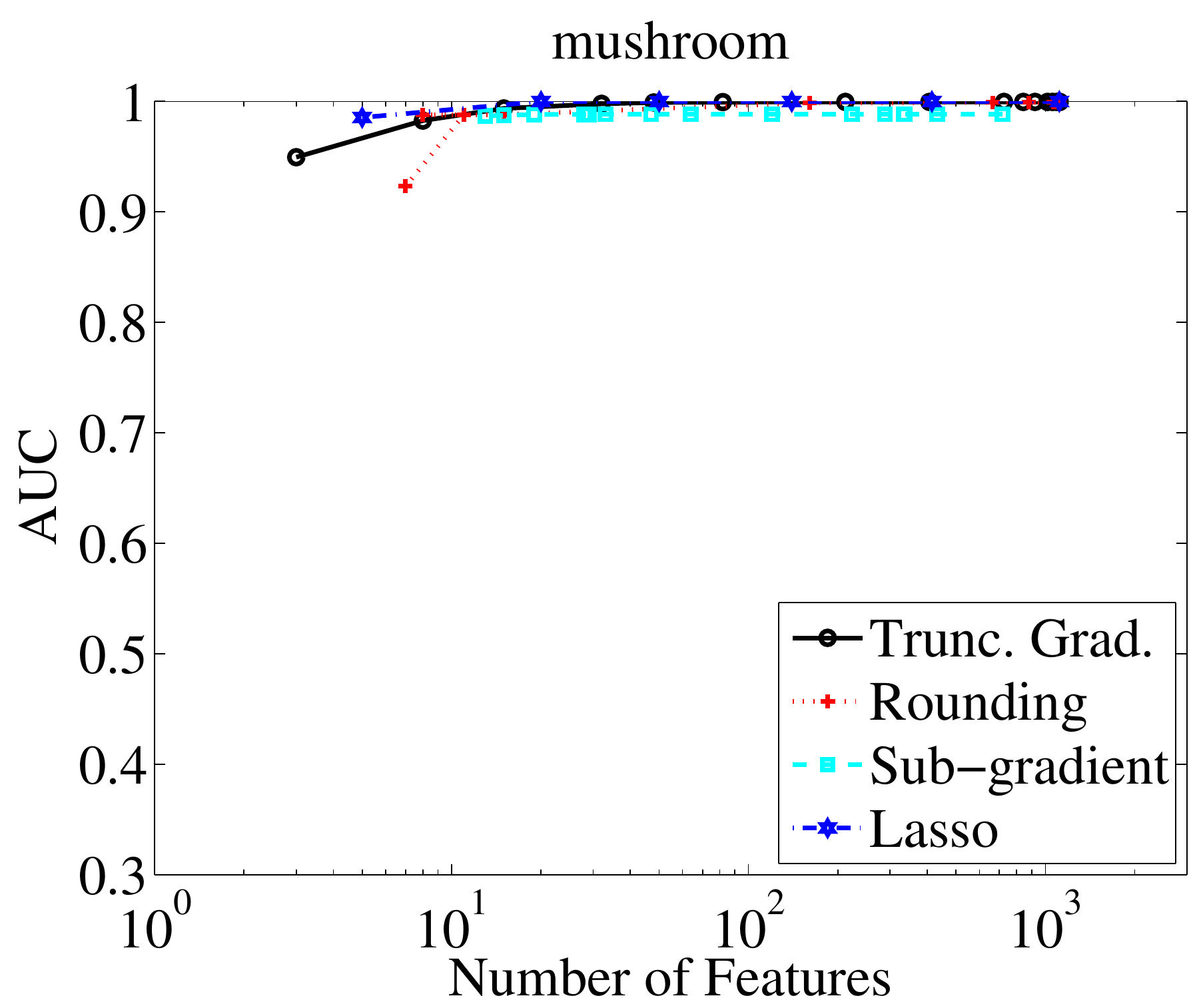}
&
\includegraphics[width=0.4\columnwidth]{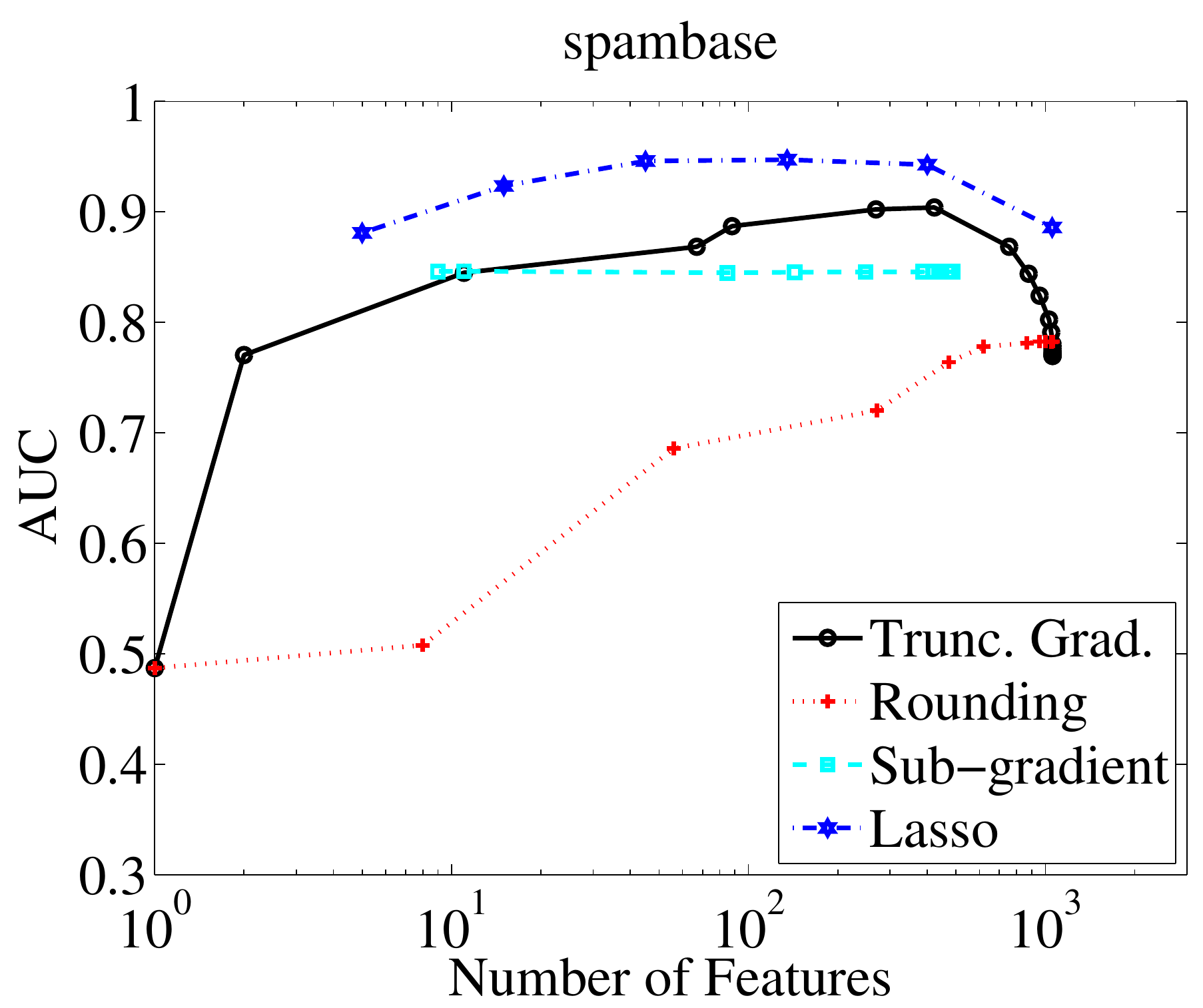}
\\
\includegraphics[width=0.4\columnwidth]{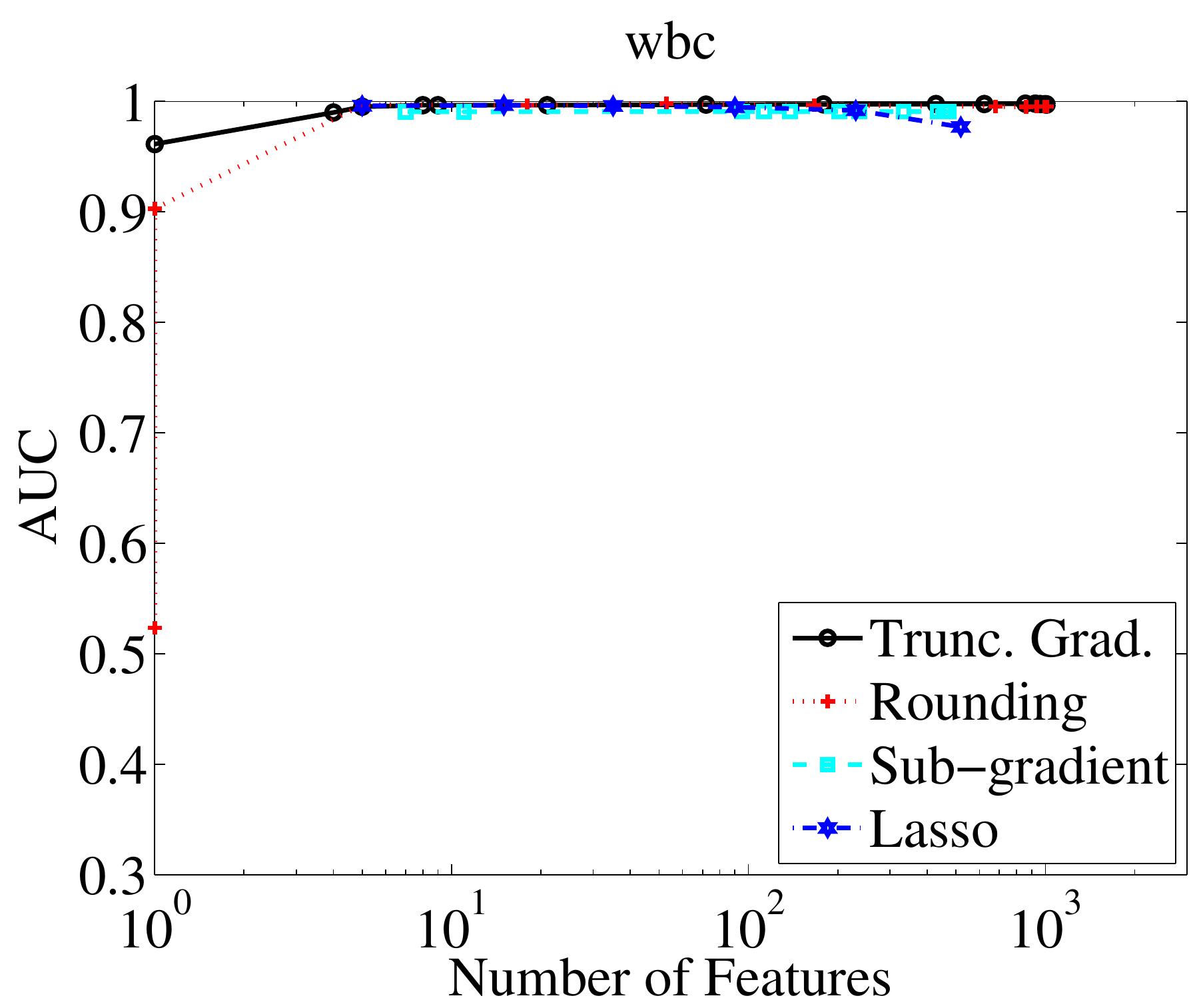}
&
\includegraphics[width=0.4\columnwidth]{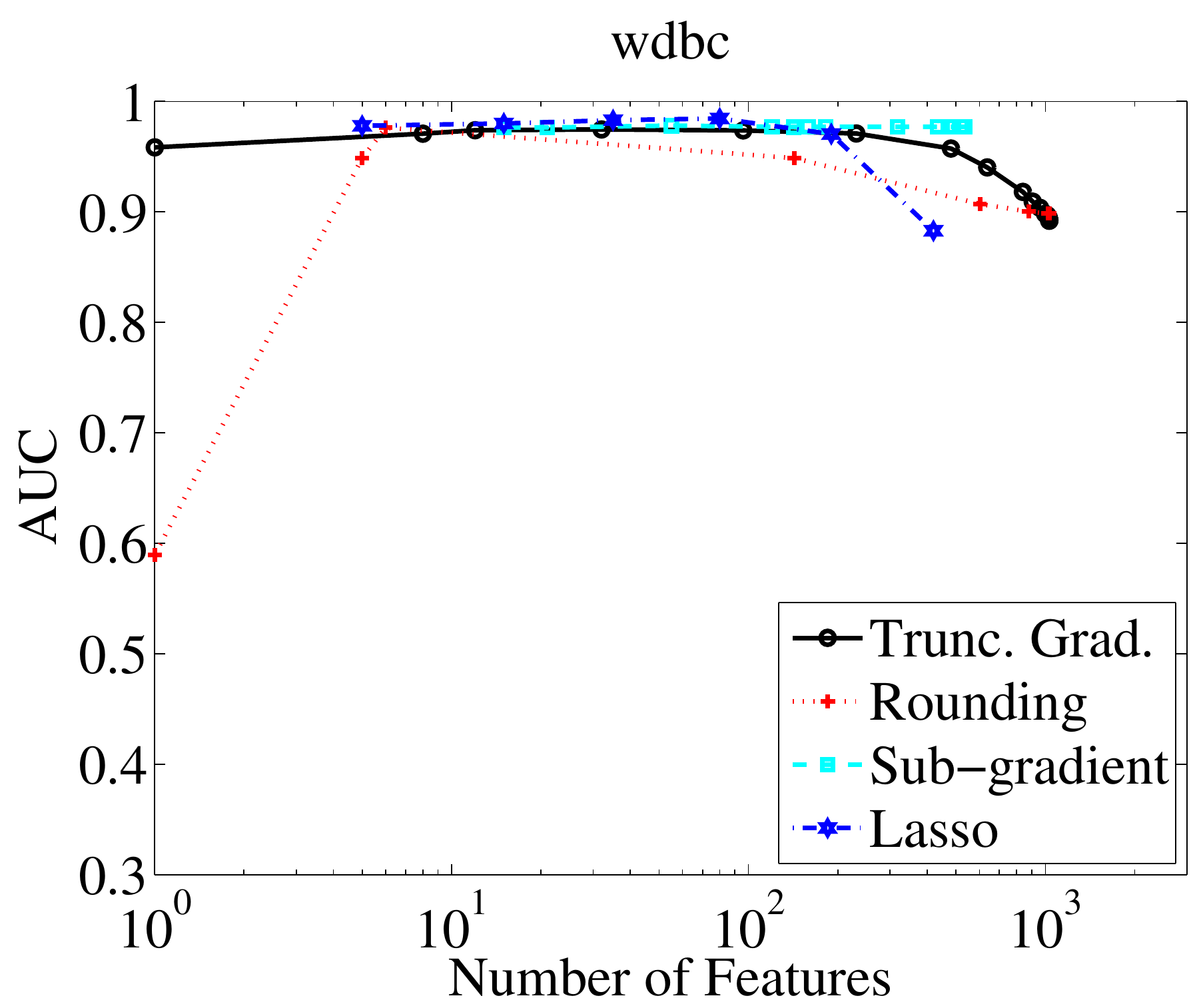}
\end{tabular}
\end{center}
\caption{Comparison of four algorithms.} \label{fig:comparisons}
\end{figure}

Figure~\ref{fig:comparisons} gives the results.  First, it is
observed that truncated gradient is consistently competitive with
the other two online algorithms and significantly outperformed
them in some problems.  This suggests the effectiveness of
truncated gradient.

Second, it is interesting to observe that the qualitative behavior
of truncated gradient is often similar to that of LASSO,
especially when very sparse weight vectors are allowed (the left
sides in the graphs).  This is consistent with
theorem~\ref{thm:sparse-online-stochastic} showing the relation
between these two algorithms.  However, LASSO usually has worse
performance when the allowed number of nonzero weights is set too
large (the right side of the graphs).  In this case, LASSO seems
to overfit.  In contrast, truncated gradient is more robust to
overfitting. The robustness of online learning is often attributed
to early stopping, which has been extensively discussed in
the literature (e.g., in \cite{Zhang04-icml}).

Finally, it is worth emphasizing that the experiments in this
subsection try to shed some light on the relative strengths of
these algorithms in terms of feature sparsification.  For large
datasets such as \bigads only truncated gradient, coefficient rounding,
and the sub-gradient algorithms are applicable to large-scale problems
with sparse features.  As we have shown and argued, the rounding algorithm
is quite ad hoc and may not work robustly in some problems, and the
sub-gradient algorithm does not lead to sparsity in general during training.

\section{Conclusion}

This paper covers the first sparsification technique for large-scale
online learning with strong theoretical guarantees.  The algorithm,
truncated gradient, is the natural extension of Lasso-style regression
to the online-learning setting.  Theorem~\ref{thm:sparse-online-regret}
proves that the technique is sound: it never harms performance much
compared to standard stochastic gradient descent in adversarial situations.
Furthermore, we show that the asymptotic solution of one instance of the
algorithm is essentially equivalent
to Lasso regression, and thus justifying the algorithm's ability to
produce sparse weight vectors when the number of features is intractably large.

The theorem is verified experimentally in a number of problems.  In some cases, especially
for problems with many irrelevant features, this approach achieves a one or
two order of magnitude reduction in the number of features.

\bibliographystyle{plain}
\bibliography{refs}

\appendix

\section{Proof of Theorem~\ref{thm:sparse-online-regret}}

The following lemma is the essential step in our analysis.
\begin{lemma} \label{lem:online-update-onestep}
For update rule (\ref{eq:sparse-online-update-truncated-grad}) applied to weight
vector $w$ on example $z=(x,y)$ with gravity parameter $g_i=g$,
resulting in a weight vector $w'$.
If Assumption~\ref{assump:loss} holds,
then for all $\bar{w} \in R^d$, we have
\begin{align*}
& (1-0.5 A\eta) L(w,z) + g \|w' \cdot I (|w'| \leq \theta)\|_1 \\
\leq& L(\bar{w},z) +  g \|\bar{w} \cdot I (|w'| \leq \theta) \|_1 + \frac{\eta}{2} B + \frac{\|\bar{w}-w\|^2 - \|\bar{w}-w'\|^2}{2\eta} .
\end{align*}
\end{lemma}
\begin{proof}
Consider any target vector $\bar{w} \in R^d$ and let
$\tilde{w}=w - \eta \nabla_1 L (w,z)$.  We have
$w'=T_1(\tilde{w}, g\eta, \theta)$.
Let
\[
u(\bar{w},w')= g \|\bar{w} \cdot I (|w'| \leq \theta) \|_1 -
 g \|w' \cdot I (|w'| \leq \theta)\|_1 .
\]
Then the update equation implies the following:
\begin{align*}
&\|\bar{w}-w'\|^2 \\
\leq & \|\bar{w}-w'\|^2 + \|w'-\tilde{w}\|^2 \\
= & \|\bar{w}-\tilde{w}\|^2 - 2(\bar{w} - w')^T(w' - \tilde{w})\\
\leq &\|\bar{w}-\tilde{w}\|^2
+ 2 \eta u(\bar{w},w') \\
=& \|\bar{w}-w\|^2 + \|w-\tilde{w}\|^2 + 2(\bar{w}-w)^T (w-\tilde{w})
 + 2 \eta u(\bar{w},w') \\
=& \|\bar{w} - w\|^2 + \eta^2 \|\nabla_1 L (w,z) \|^2
 + 2\eta (\bar{w}-w)^T \nabla_1 L (w,z) + 2 \eta u(\bar{w},w') \\
\leq& \|\bar{w}-w\|^2 + \eta^2 \|\nabla_1 L (w,z) \|^2
 + 2\eta (L (\bar{w},z)-L (w,z)) +  2 \eta u(\bar{w},w') \\
\leq& \|\bar{w}-w\|^2  + \eta^2 (A L(w,z)+B)
 + 2\eta (L (\bar{w},z)-L (w,z)) + 2 \eta u(\bar{w},w') .
\end{align*}
Here, the first and second equalities follow from algebra, and the third
from the definition of $\tilde{w}$.  The first inequality follows because
a square is always non-negative.  The second inequality follows because
$w'=T_1(\tilde{w},g\eta,\theta)$, which implies that
$(w'-\tilde{w})^T w' = - g\eta \|w'\cdot I(|w'| \leq \theta) \|_1$
and $|w'_j-\tilde{w}_j| \leq  g\eta I(|w'_j| \leq \theta)$.
Therefore
\begin{align*}
-(\bar{w} - w')^T(w' - \tilde{w})
=&-\bar{w}^T (w' - \tilde{w}) + {w'}^T (w'-\tilde{w})\\
\leq & \sum_{j=1}^d  |\bar{w}_j| |w'_j-\tilde{w}_j|
+  (w'-\tilde{w})^T w' \\
\leq&  g \eta \sum_{j=1}^d  |\bar{w}_j| I(|w'_j| \leq \theta)
+  (w'-\tilde{w})^T w'
= \eta u(\bar{w},w') .
\end{align*}
The third
inequality follows from the definition of sub-gradient of a convex function,
which implies that
\[
(\bar{w}-w)^T \nabla_1 L (w,z) \leq L (\bar{w},z)-L (w,z)
\]
for all $w$ and $\bar{w}$.
The fourth
inequality follows from Assumption~\ref{assump:loss}.
Rearranging the above inequality leads to the desired bound.
\end{proof}

\begin{proof} (of theorem \ref{thm:sparse-online-regret})
Apply Lemma~\ref{lem:online-update-onestep} to the update on trial
$i$, we have
\begin{align*}
& (1-0.5 A\eta) L(w_i,z_i) + g_i \|w_{i+1}\cdot I(|w_{i+1}| \leq \theta) \|_1 \\
\leq& L(\bar{w},z_i) + \frac{\|\bar{w}-w_i\|^2 - \|\bar{w}-w_{i+1}\|^2 }{2\eta}
 + g_i \| \bar{w}\cdot I(|w_{i+1}| \leq \theta) \|_1 + \frac{\eta}{2} B .
\end{align*}
Now summing over $i=1,2,\ldots,T$, we obtain
\begin{eqnarray*}
& &  \sum_{i=1}^T \left[ (1-0.5 A\eta)L(w_i,z_i) + g_i \|w_{i+1}\cdot I(|w_{i+1}| \leq \theta)\|_1 \right] \\
&\leq& \sum_{i=1}^T \left[ \frac{\|\bar{w}-w_i\|^2 - \|\bar{w}-w_{i+1}\|^2 }{2\eta}
 + L(\bar{w},z_i) + g_i \| \bar{w} \cdot I(|w_{i+1}| \leq \theta)\|_1 + \frac{\eta}{2} B \right] \\
&=& \frac{\|\bar{w}-w_1\|^2 - \|\bar{w}-w_{T}\|^2 }{2\eta}  + \frac{\eta}{2} T B
+ \sum_{i=1}^T [L(\bar{w},z_i) +g_i \| \bar{w} \cdot I(|w_{i+1}| \leq \theta)\|_1]
\\
&\leq & \frac{\|\bar{w}\|^2 }{2\eta} +
 \frac{\eta}{2} T B
+ \sum_{i=1}^T [L(\bar{w},z_i) +g_i \| \bar{w} \cdot I(|w_{i+1}| \leq \theta)\|_1] .
\end{eqnarray*}
The first equality follows from the telescoping sum and the second
inequality follows from the initial condition (all weights are zero)
and dropping negative quantities.
The theorem follows by dividing with respect to $T$ and rearranging terms.
\end{proof}

\end{document}